\documentclass[11pt]{now_as_preprint}

\pdfoutput=1

\usepackage{amssymb}
\usepackage{amsmath}
\allowdisplaybreaks[3] 
\usepackage{bbold} 

\usepackage{flafter} 
\PassOptionsToPackage{pdftex}{graphicx} 

\usepackage[pdftex]{graphicx}

\usepackage{fancyhdr} 
\usepackage{ifthen} 
\usepackage{layout} 
\usepackage{xspace}
\usepackage{afterpage}
\usepackage{url}
\usepackage{textcomp} 

\usepackage{wasysym} 

\usepackage[latin1]{inputenc} 

\def\ba#1\ea{\begin{align*}#1\end{align*}} 
\def\banum#1\eanum{\begin{align}#1\end{align}} 

\newcommand{\bi}{\begin{itemize}}
\newcommand{\ei}{\end{itemize}}
\newcommand{\be}{\begin{enumerate}}
\newcommand{\ee}{\end{enumerate}}
\newcommand{\bc}{\begin{center}}
\newcommand{\ec}{\end{center}}

\ifx\theorem\undefined
\newtheorem{theorem}{Lemma}

\newtheorem{conjecture}[theorem]{Conjecture}

\else
\fi

\newcommand{\ulesqed}{\hfill\smiley}

\ifx\proof\undefined
\newenvironment{proof}{\par\noindent{\em Proof. }}{\ulesqed\\[2mm]}
\else
\fi

% ...

\newcommand{\ulesquote}[1]{
\begin{list}{}{%
\setlength{\leftmargin}{0.5cm}%
\setlength{\rightmargin}{0cm}%
\setlength{\topsep}{0cm}%
}
\item[] #1
\end{list}
}

\newcommand{\RR}{\mathbb{R}}
 
\newcommand{\Nat}{\mathbb{N}}

\let\R\undefined 
\newcommand{\R}{\RR}

\newcommand{\Acal}{\mathcal{A}}

\newcommand{\Ccal}{\mathcal{C}}

\newcommand{\Xcal}{\mathcal{X}}

\DeclareMathOperator*{\argmin}{argmin} 

\DeclareMathOperator{\instab}{Instab}

\newcommand{\eps}{\ensuremath{\varepsilon}}
\renewcommand{\epsilon}{\ensuremath{\varepsilon}}
\renewcommand{\phi}{\ensuremath{\varphi}}
\newcommand{\charfct}{\mathbb{1}}

\newcommand{\union}{\cup}  

\newcommand{\hd}{\hdots}

\newcommand{\fig}[1]{Figure~\protect\ref{#1}}

\renewcommand{\sec}[1]{Section~\protect\ref{#1}}

\newcommand{\eq}[1]{Equation~\protect(\ref{#1})}

\renewcommand{\th}[1]{Theorem~\protect\ref{#1}}

\newcommand{\blobb}[1]{%
\begin{list}{$\bullet$}{%
\setlength{\topsep}{0cm}
\setlength{\leftmargin}{0.5cm}
}{\item #1}%
\end{list}
}

\wasyfamily
\DeclareFontShape{U}{wasy}{b}{n}{ <-10> ssub * wasy/m/n
<10> <10.95> <12> <14.4> <17.28> <20.74> <24.88>wasyb10 }{}

\DeclareFontShape{U}{wasy}{m}{n}{ <5> <6> <7> <8> <9> gen * wasy
<10> <10.95> <12> <14.4> <17.28> <20.74> <24.88> <35> <40> <50> <60> wasy10  }{}

\usepackage[round,comma]{natbib}
\newcommand{\dmm}{d_{\text{MM}}} %
\newcommand{\dboundary}{d_{\text{boundary}}} %
\newcommand{\rinstab}{\text{RInstab}} 

\newcommand{\cn}{c^{(n)}}
\newcommand{\cstar}{c^{(*)}}
\newcommand{\n}{^{(n)}}

\newtheorem{conclusion}[theorem]{Conclusion}

\newtheorem{conjecture}[theorem]{Conjecture}
\newcommand{\ktrue}{K_{\text{true}}}
\newcommand{\kinit}{K_{\text{init}}}

\begin{document}

\articletitle{Clustering stability: an overview}

\authorname1{Ulrike von Luxburg}
\author1email{ulrike.luxburg@tuebingen.mpg.de}

\author1address2ndline{Max Planck Institute for Biological
Cybernetics, T{\"u}bingen, Germany}


\abstract{A popular method for selecting the number of clusters is
  based on stability arguments: one chooses the
  number of clusters such that the corresponding clustering results
  are ``most
  stable''. In recent years, a series of papers has analyzed the
  behavior of this method from a theoretical point of view. However,
  the results are very technical and difficult to interpret for
  non-experts.  In this paper we give a high-level overview about the
  existing literature on clustering stability. In addition to
  presenting the results in a slightly informal but accessible way, we
  relate them to each other and discuss their different
  implications. %
}

\maketitle

\cleardoublepage 
\pagenumbering{roman}
\tableofcontents
\clearpage

\setcounter{page}{235}
\pagenumbering{arabic}

\setcounter{tocdepth}{4}

\chapter{Introduction} \label{sec-intro}

\begin{figure}[t]
\begin{center}
\includegraphics[width=0.6\textwidth]{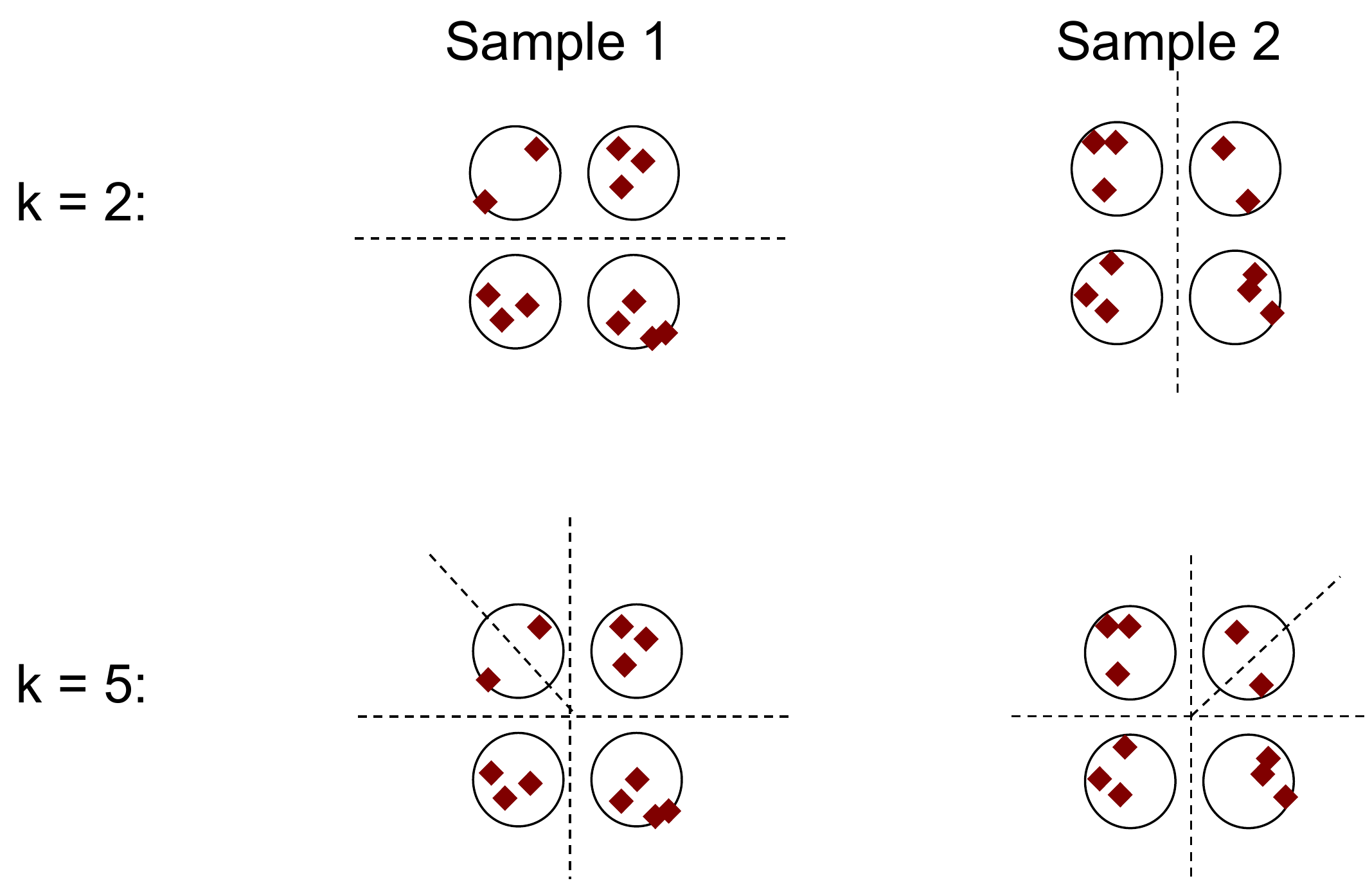}
\end{center}
\caption{Idea of clustering stability. Instable clustering solutions if the number of clusters is too small
  (first row) or too large (second row). See text for
  details. }
\label{fig-stability-idea}
\end{figure}

Model selection is a difficult problem in non-parametric
clustering. The obvious reason is that, as opposed to supervised
classification, there is no ground truth against which we could
``test'' our clustering results. One of the most pressing questions in
practice is how to determine the number of clusters. Various ad-hoc
methods have been suggested in the literature, but none of them is
entirely convincing. These methods usually suffer from the fact that
they implicitly have to define ``what a clustering is'' before they
can assign different scores to different numbers of clusters. In
recent years a new method has become increasingly popular:
selecting the number of clusters based on clustering
stability. Instead of defining ``what is a clustering'', the basic
philosophy is simply that a clustering should be a structure on the
data set that is ``stable''.  That is, if applied to several data
sets from the same underlying model or of the same data generating
process, a clustering algorithm should obtain similar results.  In this
philosophy it is not so important
how the clusters look (this is taken care of by the
clustering algorithm), but that they can be constructed
in a stable manner. \\

The basic intuition of why people believe that this
is a good principle can be described by \fig{fig-stability-idea}. 
Shown is a data distribution with four underlying clusters
(depicted by the black circles), and different samples from this
distribution (depicted by red diamonds). If we cluster this data set into $K=2$
clusters, there are two reasonable solutions: a horizontal and a
vertical split. If a clustering algorithm is applied repeatedly to
different samples from this distribution, it might sometimes construct the
horizontal and sometimes the vertical solution. Obviously, these two
solutions are very different from each other, hence the clustering
results are instable. Similar effects take place if we start with
$K=5$. In this case, we necessarily have to split an existing cluster
into two clusters, and depending on the sample this could happen to
any of the four clusters. Again the clustering solution is
instable. Finally, if we apply the algorithm with the correct number
$K=4$, we observe stable results (not shown in the figure):
the clustering algorithm always discovers the correct clusters
(maybe up to a few outlier points). In this example, the 
stability principle detects the correct number of clusters. \\

At first glance, using stability-based principles for model selection
appears to be very attractive.  It is elegant as it avoids to define what
a good clustering is. It is a meta-principle that can be applied to
any basic clustering algorithm and does not require a particular
clustering model. Finally, it sounds ``very fundamental'' from a
philosophy of inference point of view. \\

However, the longer one thinks about this principle, the less obvious
it becomes that model selection based on clustering stability ``always
works''.  What is clear is that solutions that are completely instable
should not be considered at all. However, if there are several stable
solutions, is it always the best choice to select the one
corresponding to the most stable results? One could
conjecture that the most
stable parameter always corresponds to the simplest solution, but clearly there exist
situations where the most simple solution is not what we are looking
for. To find
out how model selection based on clustering stability works we need
theoretical results. \\

In this paper we discuss a series of theoretical results on clustering
stability that have been obtained in recent years. In Section
\ref{sec-implementation} we review different protocols for how clustering
stability is computed and used for model selection. In Section
\ref{sec-kmeans} we concentrate on theoretical results for the
$K$-means algorithm and discuss their various relations. This is the
main section of the paper. Results for more general clustering
algorithms are presented in Section
\ref{sec-beyond}. \\

\chapter{Clustering stability: definition and implementation}  \label{sec-implementation}

A {\em clustering of a data set $S = \{X_1, \hd, X_n\}$} is a function that assigns labels
to all points of $S$, that is 
$
\Ccal_K: S \to \{1, \hd, K\}. 
$ 
Here $K$ denotes the number of clusters. %
A {\em clustering algorithm} is a procedure that takes a set $S$ of
points as input and outputs a clustering of $S$.
The clustering algorithms considered in this paper take an additional
parameter as input, namely the number $K$ of clusters they are
supposed to construct.

We analyze clustering stability in a {\em statistical setup}.  The
data set $S$ is assumed to consist of $n$ data points $X_1, \hdots, X_n$ that have been drawn
independently from some unknown underlying distribution $P$ on some
space $\Xcal$. The final goal is to use these sample points to
construct a good partition of the underlying space $\Xcal$.  For some
theoretical results it will be  easier to ignore sampling effects and
directly work on the underlying space $\Xcal$ endowed with the
probability distribution $P$. This can be considered as the case of
having ``infinitely many'' data points. We sometimes call this the
limit case for $n \to
\infty$. \\

Assume we agree on a way to compute distances $d(\Ccal, \Ccal')$ between different
clusterings $\Ccal$ and $\Ccal'$ (see
below for details). Then, for a fixed probability distribution $P$, a fixed number $K$ of clusters
and a fixed sample size $n$, the {\em instability of a clustering
algorithm}  is defined as the expected distance between two
clusterings $\Ccal_K(S_n)$, $\Ccal_K(S_n')$ on different data sets $S_n$, $S_n'$ of size $n$, that is 
\banum \label{def-instab}
\instab(K, n) := 
E\big( \; 
d( \Ccal_K(S_n),\Ccal_K(S_n') ) \; \big)
\eanum
The expectation is taken with respect to the drawing of the two
samples. \\

In practice, a large variety of methods has been devised to compute
stability scores and use them for model selection. On a very general
level they works as follows: \\

\ulesquote{
\small\tt
Given: a set $S$ %
of data points, a clustering
algorithm $\Acal$ that takes the number $k$ of clusters as input
\sloppy
\begin{enumerate}

\item For $k=2, \hd, k_{\max}$

\setlength{\rightmargin}{0pt}

\begin{enumerate}
\item Generate perturbed versions $S_b$ $(b = 1,\hd,b_{\max})$ of the original data set (for
  example by subsampling or adding noise, see below)

\item For $b = 1, \hd, b_{\max}$: \\
Cluster the data set $S_b$ with algorithm $\Acal$
  into $k$ clusters to obtain clustering $\Ccal_b$
\item For $b , b' = 1, \hd, b_{\max}$: \\
Compute pairwise distances $d(\Ccal_{b}, \Ccal_{b'})$ between
  these clusterings (using one of the distance functions described
  below)
\item Compute instability as the mean distance between clusterings~$\Ccal_b$: 
\ba
\widehat{\instab}(k,n) =  \frac{1}{b_{\max}^2}
 \sum_{b, b' =1}^{b_{max}} \; d(\Ccal_{b}, \Ccal_{b'})
\ea
\end{enumerate}

\item Choose the parameter $k$ that gives the best stability, in
  the simplest case as follows:
\ba
K := \argmin_k \widehat{\instab}(k,n)
\ea
(see below for more options). 
\end{enumerate} 
}

This scheme gives a very rough overview of how clustering stability can
be used for model selection. In practice, many details have to be
taken into account, and they will be discussed in the next
section. Finally, we want to mention an approach that is vaguely
related to clustering stability, namely the ensemble method
\citep{StrGho02}. Here, an
ensemble of {\em algorithms} is applied to one fixed data set. Then a
final clustering is built from the results of the individual
algorithms. We are not going to discuss this approach in our paper. \\

{\bf Generating perturbed versions of the data set. } 
To be able to evaluate the stability of a fixed clustering algorithm
we need to run the clustering algorithm several times on slightly
different data sets. To this end we need to generate perturbed versions
of the original data set. In practice, the following schemes have
been used:

\begin{itemize}
\item Draw a random subsample of the original data set without
  replacement 
\citep{LevDom01,BenEliGuy02,FriDud01,LanRotBraBuh04}. \\

\item Add random noise to the original data
points \citep{Bittner00_long,MolRad06}. \\

\item If the original data set is high-dimensional, use different random
projections in low-dimensional spaces, and then cluster the
low-dimensional data sets \citep{SmoGho03}. \\

\item If we work in a model-based framework, sample data from the 
model \citep{KerChu01}. \\

\item Draw a random sample of the original data with
  replacement. This approach has not been reported in the literature
  yet, but it avoids the problem of setting the size of the
  subsample. For good reasons, this kind of sampling is the standard in the bootstrap
  literature \citep{EfrTib93} and might also have advantages in the stability
  setting. This scheme requires that the algorithm
  can deal with weighted data points (because some data points will
  occur several times in the sample). \\

\end{itemize}

In all cases, there is a trade-off that has to be treated
carefully. If we change the data set too much (for example, the
subsample is too small, or the noise too large), then we might destroy
the structure we want to discover by clustering. If we change the data
set too little, then the clustering algorithm will always obtain the
same results, and we will observe trivial stability. It is hard to
quantify this trade-off in practice. \\

{\bf Which clusterings to compare? }
Different protocols are used to compare the clusterings on the
different data sets $S_b$. 

\begin{itemize} 

\item Compare the clustering of the original data set with the
  clusterings obtained on subsamples \citep{LevDom01}. \\

\item  Compare clusterings of overlapping subsamples on the data
  points where both clusterings are defined. 
  \citep{BenEliGuy02}. \\

\item Compare clusterings of disjoint subsamples 
\citep{FriDud01, LanRotBraBuh04}. Here we first need to apply an
extension operator to extend each clustering to the domain of the other
clustering. \\

\end{itemize}

{\bf Distances between clusterings. } If two clusterings are defined
on the same data points, then it is straightforward to compute a
distance score between these clusterings based on any of the
well-known clustering distances such as the Rand index, Jaccard index,
Hamming distance, minimal matching distance, Variation of Information
distance \citep{Meila03_colt}. All these distances count, in some way or
the other, points or pairs of points on which the two clusterings
agree or disagree. The most convenient choice from a theoretical point
of view is the minimal matching distance. For two clusterings $\Ccal,
\Ccal'$ of the same data set of $n$ points it is defined as 
\banum \label{def-dmm}
\dmm(\Ccal, \Ccal') := 
\min_\pi 
\frac{1}{n}\sum_{i=1}^n \charfct_{ \{\Ccal(X_i) \neq \pi( \Ccal'(X_i))   \}}
\eanum
where the minimum is taken over all permutations $\pi$ of the $K$
labels. Intuitively, the minimal matching distance measures the same
quantity as the 0-1-loss used in supervised classification. For a
stability study involving the adjusted Rand index or an adjusted
mutual information index see \citet{VinEpp09}. \\ 

If two clusterings are defined on different data sets one has two
choices. If the two data sets have a big overlap one can use a {\em
  restriction operator} to restrict the clusterings to the points
that are contained in both data sets. On this restricted set one can
then compute a standard distance between the two clusterings. The
other possibility is to use an
{\em extension operator} to extend both clusterings from their domain
to the domain of the other clustering. Then one can compute a standard
distance between the two clusterings as they are now both defined on the
joint domain.  For center-based clusterings, as constructed by the
$K$-means algorithm, a natural extension operator exists. Namely, to a
new data point we simply assign the label of the closest cluster
center. A more general scheme to extend an existing clustering to new
data points is to train a classifier on the old data points and use
its predictions as labels on the new data points. However, in the
context of clustering stability it is not obvious what kind of bias we
introduce with this approach. \\

\begin{figure}[t]
\begin{center}
\includegraphics[width=0.49\textwidth]{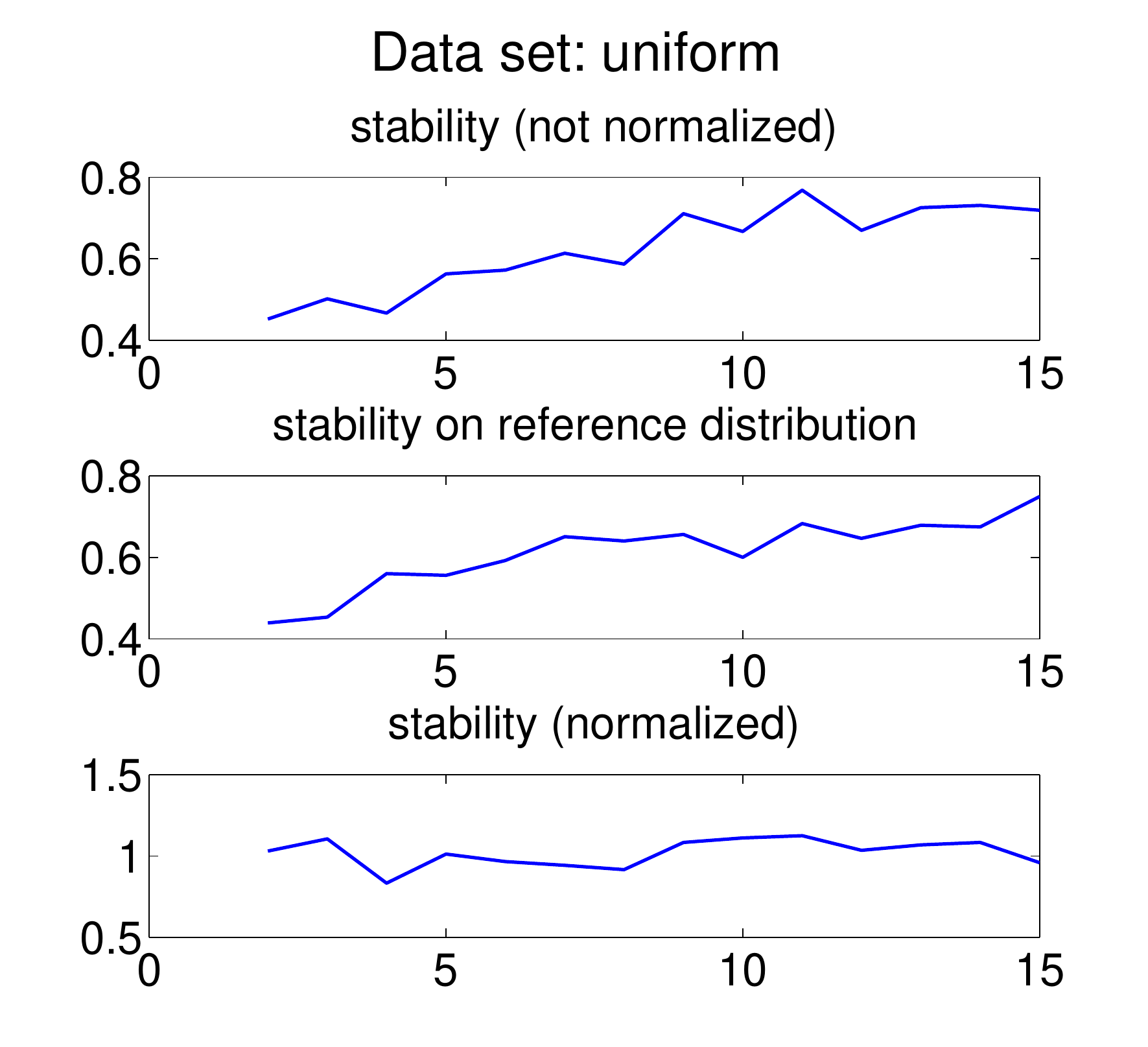}\hfill
\includegraphics[width=0.49\textwidth]{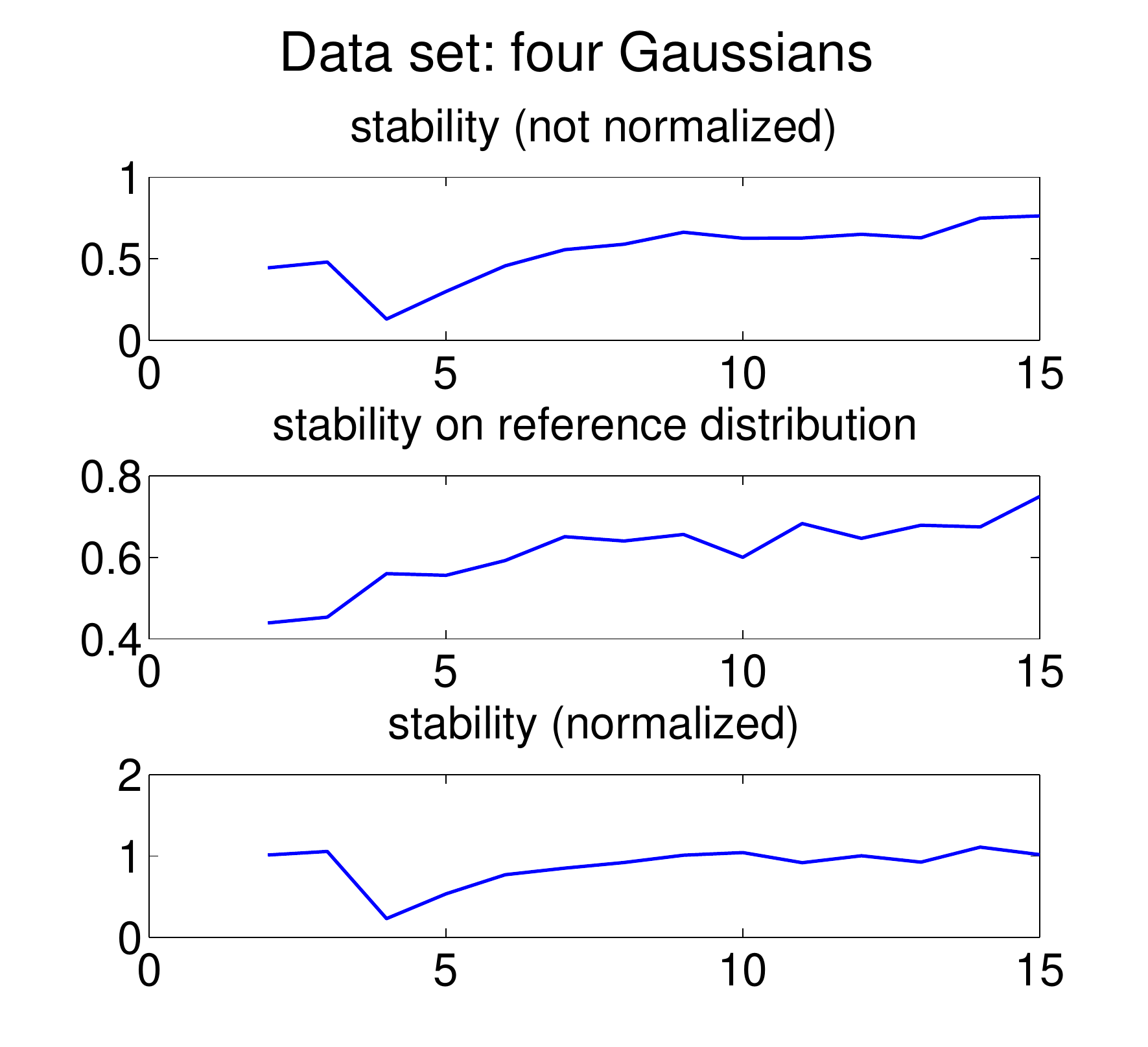}
\end{center}
\caption{Normalized stability scores. Left plots: data points from a
  uniform density on $[0,1]^2$. Right plots: data points from a
  mixture of four well-separated Gaussians in $\R^2$. The first row
  always shows the unnormalized instability $\widehat{\instab}$ for $K=2, ...,
  15$. The second row shows the instability $\widehat{\instab}_{\text{norm}}$
  obtained on a reference distribution (uniform distribution). The
  third row shows the normalized stability $\widehat{\instab}_{\text{norm}}$.}
\label{fig-stability-normalized}
\end{figure}

{\bf Stability scores and their normalization. }
The stability protocol outlined above results in a set of distance
values $(d(\Ccal_{b}, \Ccal_{b'}))_{b,b' = 1, \hd, b_{max}}$. In most approaches, one summarizes
these values by taking their mean: 
\ba
\widehat{\instab}(k,n) =  \frac{1}{b_{\max}^2}
 \sum_{b, b' =1}^{b_{max}} \; d(\Ccal_{b}, \Ccal_{b'})
\ea
Note that the mean is the simplest summary statistic one can compute based on the distance
values $d(\Ccal_{b}, \Ccal_{b'})$. A different approach is to use the  area under the cumulative
distribution function of the distance values as the stability score, see
\citet{BenEliGuy02} or \citet{BerVal07} for details. In principle one could also
come up with more elaborate statistics based on distance values. 
To the best of our knowledge, such concepts
have not been used so far. \\

The simplest way to select the number $K$
of clusters  is to minimize  the instability:  
\ba
K = \argmin_{k=2,\hd,k_{\max}} \widehat{\instab}(k,n). 
\ea
This approach has been suggested in \citet{LevDom01}. However, an
important fact to note is that $\widehat{\instab}(k,n)$ trivially scales
with $k$, regardless of  what the underlying data structure is. For
example, in the top left  plot in
Figure~\ref{fig-stability-normalized} we can see that even
for a completely unclustered data set, $\widehat{\instab}(n,k)$
increases with $k$. 
When using stability for model selection, one should correct for
the trivial scaling of $\widehat{\instab}$, otherwise it might be
meaningless to take the minimum afterwards. There exist several different {\em normalization}
protocols: 

\begin{itemize}

\item Normalization using a reference null distribution
  \citep{FriDud01,BerVal07}. One repeatedly samples data sets from
  some reference null distribution. Such a distribution is defined on
  the same domain as the data points, but does not possess any cluster
  structure. In simple cases one can use the uniform distribution on
  the data domain as null distribution. A more practical approach is
  to scramble the individual dimensions of the existing data points
  and use the ``scrambled points'' as null distribution (see
   \citealp{FriDud01,BerVal07} for details). Once we have drawn several data sets from
  the null distribution, we cluster them using our clustering
  algorithm and compute the corresponding stability score
  $\widehat{\instab}_{\text{null}}$ as above. The {\em normalized stability} is then
  defined as $\widehat{\instab}_{\text{norm}} := \widehat{\instab} /
  \widehat{\instab}_{\text{null}}$.\\

\item Normalization by random labels \citep{LanRotBraBuh04}. First, we
  cluster each of the data sets $S_b$ as in the protocol above to
  obtain the clusterings $\Ccal_b$. Then, we randomly permute these
  labels. That is, we assign the label to data point $X_i$  that
  belonged to $X_{\pi(i)}$, where $\pi$ is a permutation of
  $\{1,\hd,n\}$. This leads to a permuted clustering $\Ccal_{b, \text{
      perm}}$. We then compute the stability score $\widehat{\instab}$ as above,
  and similarly we compute $\widehat{\instab}_{\text{perm}}$ for the permuted
  clusterings. The {\em normalized stability} is then defined as
  $\widehat{\instab}_{\text{norm}} := \widehat{\instab} / \widehat{\instab}_{\text{perm}}$.\\

\end{itemize}

Once we computed the normalized stability scores $\widehat{\instab}_{\text{norm}}$ we
can choose the number of clusters that has smallest normalized instability, that
is 
\ba
K = \argmin_{k=2,\hd,k_{\max}} \widehat{\instab}_{\text{norm}}(k,n)
\ea
This approach has been taken for example in \citet{BenEliGuy02,LanRotBraBuh04}. \\

{\bf Selecting $K$ based on statistical tests. }
A second approach to select the final number of clusters is to use
a statistical test. Similarly to the normalization
considered above, the idea is to compute stability scores not only on the
actual data set, but also on ``null data sets'' drawn from some
reference null distribution. Then one tests whether,
for a given parameter $k$, 
the stability on the actual data is significantly larger than the one
computed on the null data. If there are several values $k$ for which this
is the case, then one selects the one that is most
significant. The most well-known implementation of such a procedure
uses bootstrap methods \citep{FriDud01}. Other authors use a
$\chi^2$-test 
\citep{BerVal07}
or a test based on the Bernstein inequality
\citep{BerVal08}. \\

To summarize, there are many different implementations for selecting the
number $K$ of clusters based on stability scores. Until now, there
does not exist any convincing empirical study that thoroughly
compares all these approaches on a variety of data sets. In my
opinion, even fundamental issues such as the normalization have not
been investigated in enough detail. For example, in my experience
normalization often has no effect  whatsoever 
(but I did not conduct a thorough study either). To put stability-based model selection
on a firm ground it would be crucial to compare the different
approaches with each other in an extensive case study. \\

\chapter{Stability analysis of the $K$-means algorithm}  \label{sec-kmeans}

The vast majority of papers about clustering stability use the
$K$-means algorithm as basic clustering algorithm.  In this section we
discuss the stability results for the $K$-means algorithm in depth. Later, in Section \ref{sec-beyond} we will see how these results can be
extended to other clustering algorithms. \\

For simpler reference we briefly recapitulate the $K$-means algorithm
(details can be found in many text books, for example
\citealp{HasTibFri01}).  Given a set of $n$ data points $X_1, \hdots,
X_n \in \R^d$ and a fixed number $K$ of clusters to construct, the
$K$-means algorithm attempts to minimize the clustering objective
function
\banum \label{eq-kmeans-objective-finite}
Q\n_K(c_1, \hdots, c_K) = \frac{1}{n} 
\sum_{i=1}^n 
\min_{k=1,..,K} 
\|X_i - \text{c}_k\|^2
\eanum
where $c_1, \hdots, c_K$ denote the centers of the $K$ clusters. 
In the limit $n \to \infty$, the $K$-means clustering
is the one that minimizes the  limit objective
function
\banum \label{eq-kmeans-objective-limit}
Q^{(\infty)}_K(c_1, \hdots, c_K) = 
\int 
\min_{k=1,..,K} 
\|X - \text{c}_k\|^2 \; dP(X)
\eanum
where $P$ is the underlying probability distribution. \\

Given an
initial set $c^{<0>} = \{c_1^{<0>}, \hdots, c_K^{<0>}\}$ of centers, the $K$-means
algorithm iterates the following two steps until convergence: \\

\ulesquote{\small\tt
\begin{enumerate}
\item Assign data points to closest cluster centers: 
\ba
\forall i=1, \hdots, n: \;\;\; \Ccal^{<t>}(X_i) := \argmin_{k = 1, \hdots K} \| X_i - c_k^{<t>} \|
\ea

\item Re-adjust cluster means: 
\ba
\forall k=1, \hdots, K: \;\;\; 
c_k^{<t+1>}  : = 
\frac{1}{N_k} 
\sum_{ \{i \;|\;  \Ccal^{<t>}(X_i) = k\} }
X_i
\ea
where $N_k$ denotes the number of points in cluster $k$. \\
\end{enumerate}
}

It is well known that, in general, the $K$-means algorithm terminates in a
local optimum of $Q\n_K$ and does not necessarily find the global
optimum. We study the $K$-means algorithm in two
different scenarios: \\

{\bf The idealized scenario: } Here we assume an idealized algorithm
that always finds the {\em global} optimum of the $K$-means objective
function $Q\n_K$. For simplicity, we
call this algorithm the idealized $K$-means algorithm. \\

{\bf The realistic scenario: } Here we analyze the actual $K$-means
algorithm as described above. In particular, we take into account its
property of getting stuck in local optima. We also take into
account the initialization of the algorithm.\\

Our theoretical investigations are based on the following simple protocol to compute the
stability of the $K$-means algorithm: 

\be
\item We assume to have access to as many samples of size $n$ of the
  underlying distribution as we want. That is, we ignore artifacts
  introduced by computing stability on artificial perturbations of a
  fixed, given sample. 
\item As distance between two $K$-means clusterings of two samples
  $S$, $S'$ we use the minimal matching distance between the extended
  clusterings on the domain $S \union S'$. 
\item We work with the expected
  minimal matching distance as in Equation \ref{def-instab}, that
  is we analyze  $\instab$ rather than the practically
  used $\widehat{\instab}$.  This does not do
  much harm as instability scores are highly concentrated around their
  means anyway. %
\ee

\section{The idealized $K$-means algorithm}  \label{sec-idealized}

In this section we focus on the idealized $K$-means algorithm, that is
the algorithm that always finds the global optimum $c^{(n)}$ of the $K$-means
objective function:  
\ba 
\cn := (\cn_1, \hd, \cn_K) \; := \;\argmin_c \; Q\n_K(c). 
\ea

\subsection{First convergence result and the role of symmetry}  \label{sec-convergence-simple}

\begin{figure}[t]
\begin{center}
a. \includegraphics[width=0.2\textwidth]{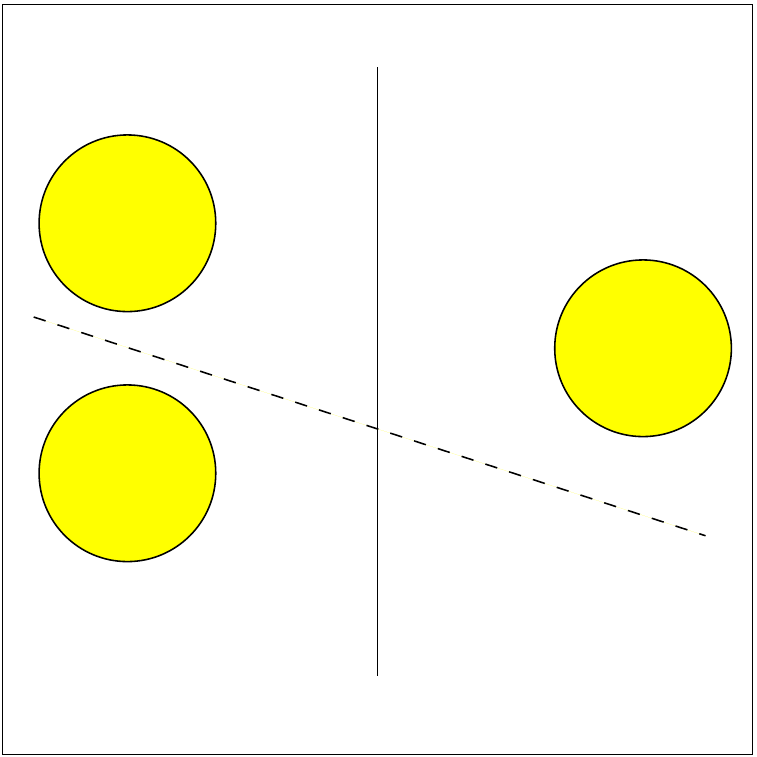}\hfill
b. \includegraphics[width=0.2\textwidth]{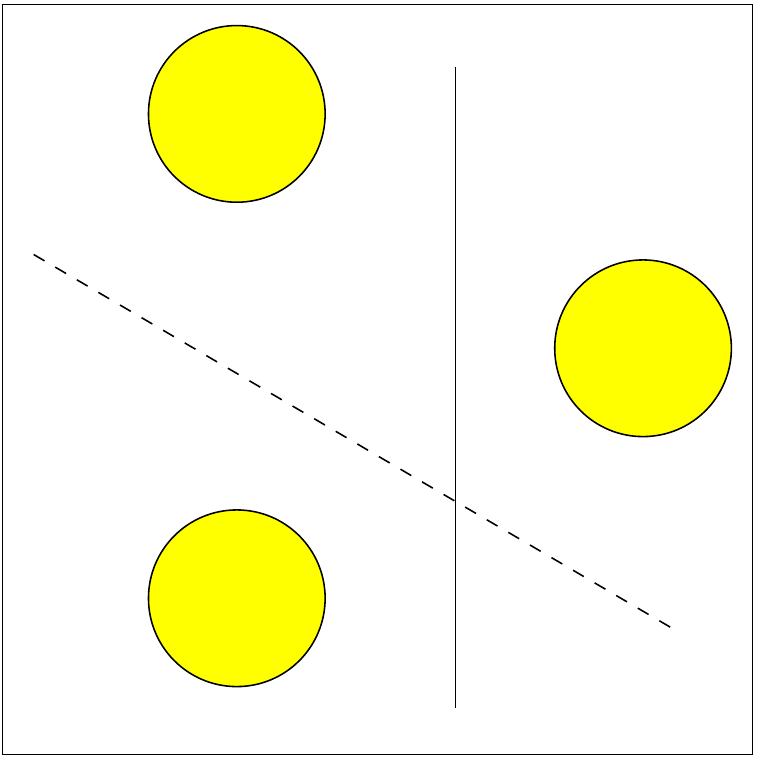}\hfill
c. \includegraphics[width=0.2\textwidth]{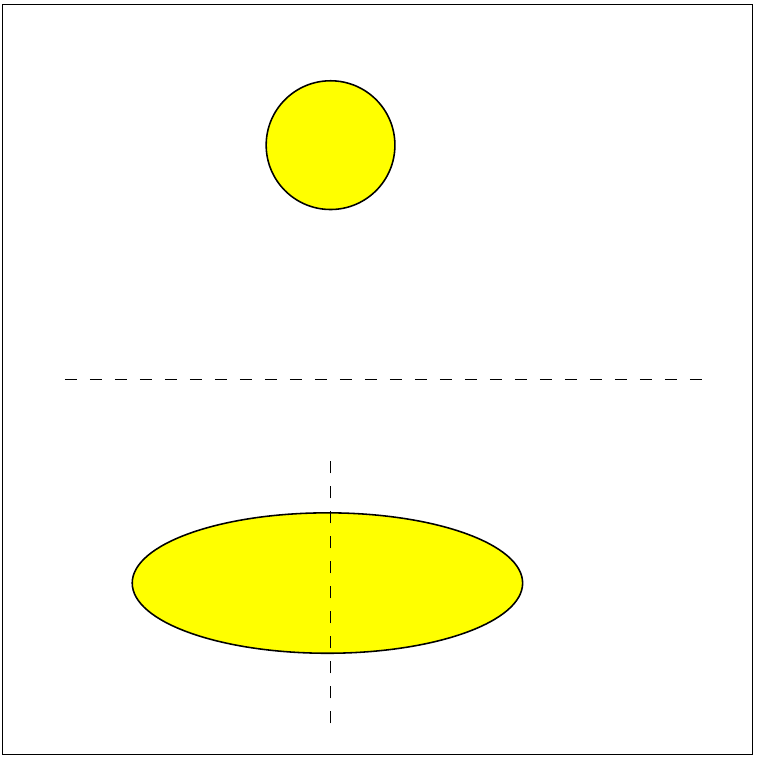}\hfill
d. \includegraphics[width=0.2\textwidth]{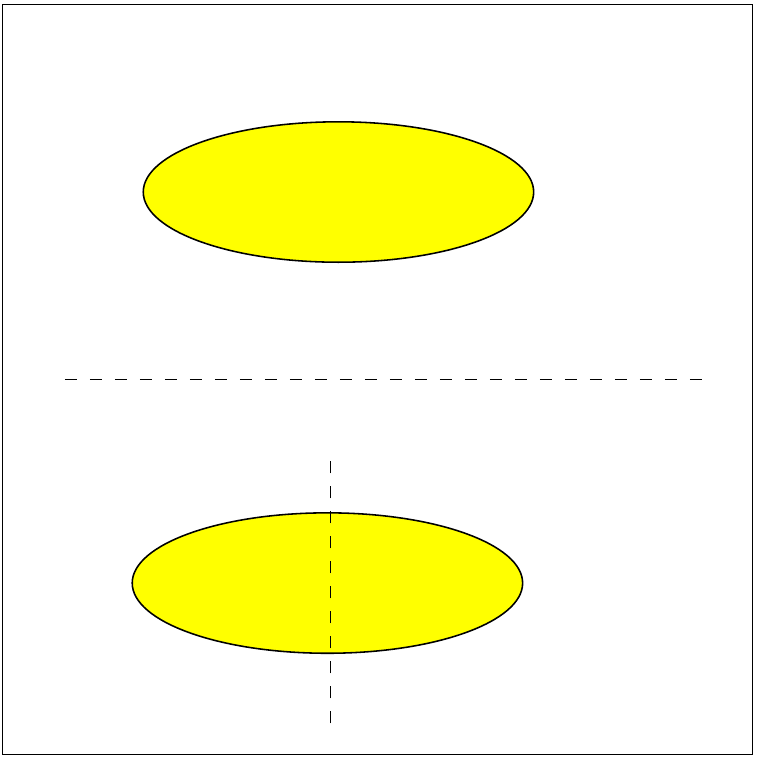}
\end{center}
\caption{If data sets are not symmetric, idealized
  $K$-means is stable even if the number $K$ of clusters is too small
  (Figure a) or too large (Figure c). Instability of the wrong number
  of clusters only occurs in symmetric data sets (Figures b and d). }
\label{fig-stability-wrong}
\end{figure}

The starting point for the results in this section is the following
observation \citep{BenLuxPal06}. %
Consider the situation in \fig{fig-stability-wrong}a. Here the data
contains three clusters, but two of them are closer to each other than
to the third cluster. Assume we run the idealized $K$-means algorithm
with $K=2$ on such a data set. Separating the left two clusters from
the right cluster (solid line) leads to a much better value of $Q\n_K$
than, say, separating the top two clusters from the bottom one (dashed
line). Hence, as soon as we have a reasonable amount of data, 
idealized (!) $K$-means with $K=2$ always constructs the first
solution (solid line). Consequently, it is stable in spite of the fact
that $K=2$ is the wrong number of clusters. Note that this would not
happen if the data set was  symmetric, as depicted in
\fig{fig-stability-wrong}b. Here neither the solution depicted by the
dashed line nor the one with the solid line is clearly superior, which
leads to instability if the idealized $K$-means algorithm is applied to
different samples. Similar examples can be constructed to detect that 
$K$ is too large, see \fig{fig-stability-wrong}c and d. With
$K=3$ it is clearly the best solution to split the big cluster in
\fig{fig-stability-wrong}c, thus clustering this data set is
stable. In \fig{fig-stability-wrong}d, however, due to symmetry
reasons neither splitting the top nor the bottom cluster
leads to a clear advantage. Again this leads to instability.\\

These informal observations suggest that unless the data set contains
perfect symmetries, the idealized $K$-means algorithm is stable even
if $K$ is wrong. This can be formalized with the following theorem. 

\begin{theorem}[Stability and global optima of the objective function]  
\label{th-convergence-simple}
Let $P$ be a
  probability distribution on $\R^d$ and $Q^{(\infty)}_K$ the limit $K$-means objective
  function as defined in \eq{eq-kmeans-objective-limit}, for some
  fixed value $K > 1$. 

\begin{enumerate}
\item If $Q^{(\infty)}_K$ has a unique global minimum, then the idealized $K$-means
  algorithm is perfectly stable when $n \to \infty$, that is 
\ba
\lim_{n \to \infty} \instab(K,n) = 0. 
\ea

\item If $Q^{(\infty)}_K$ has several global minima (for example, because the probability
  distribution is symmetric), then the idealized $K$-means
  algorithm is instable, that is
\ba
\lim_{n \to \infty} \instab(K,n) > 0. 
\ea

\end{enumerate}
\end{theorem}

This theorem has been proved (in a slightly more general setting) in 
\citet{BenLuxPal06} and \citet{BenPalSim07}. \\

{\em Proof sketch, Part 1. } It is well known that if the
objective function $Q^{(\infty)}_K$ has a unique global minimum, then the centers
$\cn$ 
constructed by the idealized $K$-means algorithm on a sample of $n$ points almost surely
converge to the true population centers $\cstar$ as $n \to
\infty$  \citep{Pollard81}. This means that given some $\eps > 0$ we can find some large
$n$ such that $\cn$ is $\eps$-close to $\cstar$ with high probability. As a
consequence, if we compare two clusterings on different samples of
size $n$, the
centers of the two clusterings are at most $2\eps$-close to each
other. Finally, one can show that if the cluster centers of two
clusterings are $\eps$-close, then their minimal matching distance is small
as well. Thus, the expected distance between the clusterings
constructed on two samples of size $n$ becomes arbitrarily small and
eventually converges to 0 as $n \to \infty$. \\
{\em Part 2.}  For simplicity, consider the symmetric situation in
\fig{fig-stability-wrong}a. Here the probability distribution has
three axes of symmetry. For $K=2$ the objective function $Q^{(\infty)}_2$ has three
global minima $c^{(*1)},c^{(*2)},c^{(*3)}$  corresponding to the three
symmetric solutions. In such a situation, the idealized $K$-means
algorithm on a sample of $n$ points gets arbitrarily close to one of
the global optima, that is $\min_{i=1, \hd, 3} d(\cn, c^{(*i)}) \to 0$
\citep{Lember03}. In particular, the sequence $(\cn)_n$ of empirical
centers has three convergent subsequences, each of which converge to
one of the global solutions. One can easily conclude that if we
compare two clusterings on random samples, with probability 1/3 they
belong to ``the same subsequence'' and thus their distance will become 
arbitrarily small. With probability 2/3 they ``belong to different
subsequences'', and thus their distance remains larger than a constant
$a >0$. %
From the latter we can conclude that
$\instab(K,n)$ is always larger than $2a/3$. 
\ulesqed \\

The interpretation of this theorem is distressing. The
stability or instability of parameter $K$ does not depend on whether
$K$ is ``correct'' or ``wrong'', but only on whether the $K$-means
objective function for this particular value $K$ has one or several
global minima. However, the number of global minima is usually not
related to the number of clusters, but rather to the fact that
the underlying probability distribution has symmetries.  In
particular, if we consider ``natural'' data distributions, such
distributions are rarely perfectly symmetric. Consequently, the
corresponding functions $Q^{(\infty)}_K$ 
usually only have one global minimum, for any value
of $K$. In practice this means that for a large sample size $n$, the
idealized $K$-means algorithm {\em is stable for any value of
$K$}. This seems to suggest that model selection based on clustering
stability does not work. However, we will see later in
\sec{sec-relationships} that this result is essentially an artifact of
the idealized clustering setting and does not carry over to the
realistic setting. \\

\subsection{Refined convergence results for the case of a unique
  global minimum} \label{sec-convergence-refined}

Above we have seen that if, for a particular distribution $P$ and a
particular value $K$, the objective function $Q^{(\infty)}_K$ has a unique global
minimum, then the idealized $K$-means algorithm is stable in the sense
that $\lim_{n \to \infty} \instab(K,n) = 0$. At first glance, this
seems to suggest that stability cannot distinguish between 
different values $k_1$ and $k_2$ (at least for large $n$). However,
this point of view is too simplistic.  It can happen that
even though both $\instab(k_1, n)$ and $\instab(k_2, n)$ converge to 0
as $n \to \infty$, this happens ``faster'' for $k_1$ than for $k_2$. 
If measured relative to the absolute
values of $\instab(k_1,n)$ and $\instab(k_2,n)$, the difference
between $\instab(k_1, n)$ and $\instab(k_2,n)$ can still be large
enough to be
``significant''. \\

The key in verifying this intuition is to study the limit process more
closely.  This line of work has been established by Shamir and Tishby
in a series of papers
\citep{ShaTis08_colt,ShaTis08_nips,ShaTis09_nips}. The main idea is
that instead of studying the convergence of $\instab(k,n)$ one needs
to consider the rescaled instability $\sqrt{n} \cdot \instab(k,
n)$. One can prove that the rescaled instability converges in
distribution, and the limit distribution depends on $k$. In
particular, the means of the limit distributions are different for
different values of $k$.  This can be formalized as follows.

\begin{theorem}[Convergence of rescaled stability]
\label{th-convergence-refined}
  Assume that the probability distribution $P$ has a density
  $p$. Consider a fixed parameter $K$, and assume that the
  corresponding limit objective function $Q^{(\infty)}_K$ has a unique global
  minimum $\cstar = (\cstar_1,\hd, \cstar_K)$.  The boundary between clusters $i$ and $j$ is
  denoted by $B_{ij}$.  Let $m \in \Nat$, and $S_{n,1}, \hd, S_{n,
    2m}$ be samples of size $n$ drawn independently from $P$. Let
  $\Ccal_K(S_{n,i})$ be the result of the idealized $K$-means
  clustering on sample $S_{n,i}$. Compute the instability as mean
  distance between clusterings of disjoint pairs of samples, that is
\banum \label{def-instab-ohad}
\overline{\instab}(K,n) :=
  \frac{1}{m} \sum_{i=1}^m \dmm \big(\Ccal_K(S_{n,2i-1}),
  \Ccal_K(S_{n,2i}) \big). 
\eanum 
Then, as $n \to \infty$ and $m \to \infty$,
  the rescaled instability $\sqrt{n} \cdot \overline{\instab}(K,n)$
  converges in probability to 
\banum \label{eq-rinstab}
\rinstab(K) := 
\sum_{1 \leq i < j \leq K} \; 
\int_{B_{ij}} \; 
\frac{ V_{ij}}{  \| \cstar_i - \cstar_j \| } \; p(x) dx, 
\eanum
where $V_{ij}$ stands for a term describing the
asymptotics of the random fluctuations of the cluster boundary between
cluster $i$ and cluster $j$ (exact formula given in 
\citealp{ShaTis08_colt,ShaTis09_nips}). 
\end{theorem}

\begin{figure}[t]
\begin{center}
\fbox{\parbox[t]{\textwidth}{
\scriptsize 
\begin{minipage}{0.15\textwidth}
\mbox{}
\end{minipage}
\begin{minipage}{0.29\textwidth}
\scriptsize 
distribution of $d(\Ccal, \Ccal')$\\
\end{minipage}
\begin{minipage}{0.2\textwidth}
\mbox{}
\end{minipage}
\begin{minipage}{0.29\textwidth}
\scriptsize  
distribution of  $\sqrt{n} \cdot d(\Ccal, \Ccal')$\\
\end{minipage}
\begin{minipage}{0.15\textwidth}
\scriptsize 
$k$ fixed \\
$n=10^2$: 
\end{minipage}
\begin{minipage}{0.29\textwidth}
\includegraphics[width=0.7\textwidth,height=0.5\textwidth]{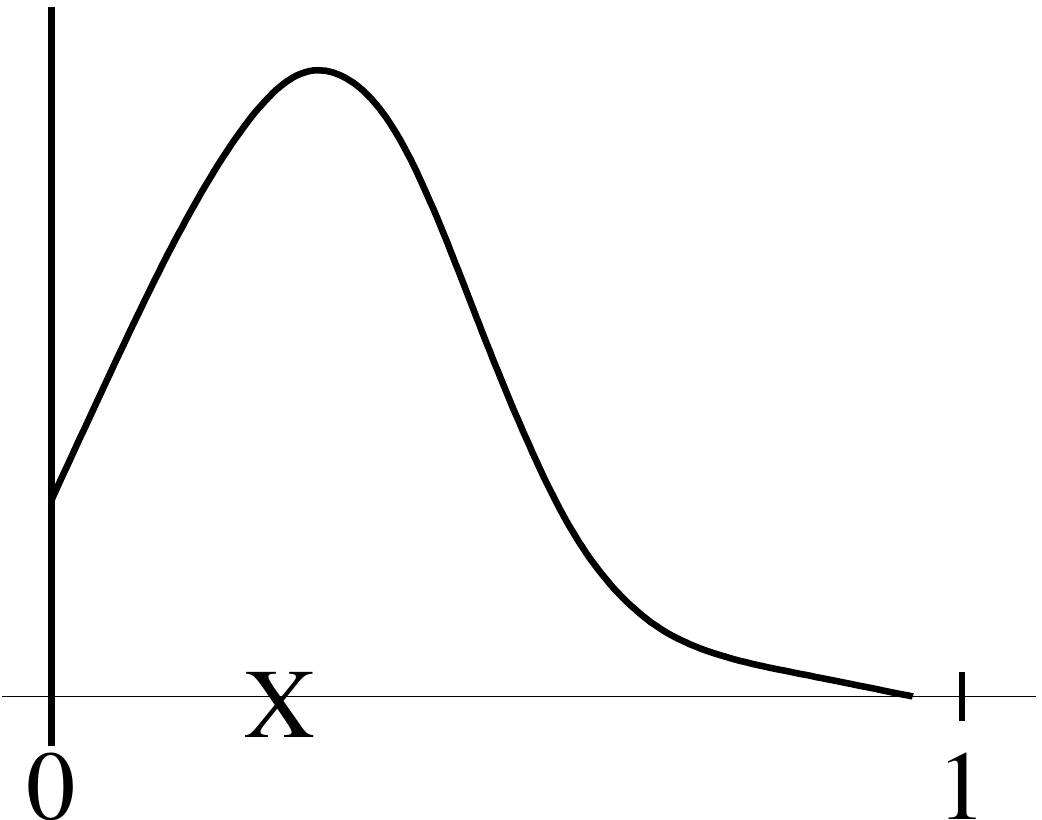}\\
\end{minipage}
\begin{minipage}{0.24\textwidth}
\small$\xrightarrow{\text{scale with } \sqrt{n}=10}$ 
\end{minipage}
\begin{minipage}{0.29\textwidth}
\includegraphics[width=0.7\textwidth,height=0.5\textwidth]{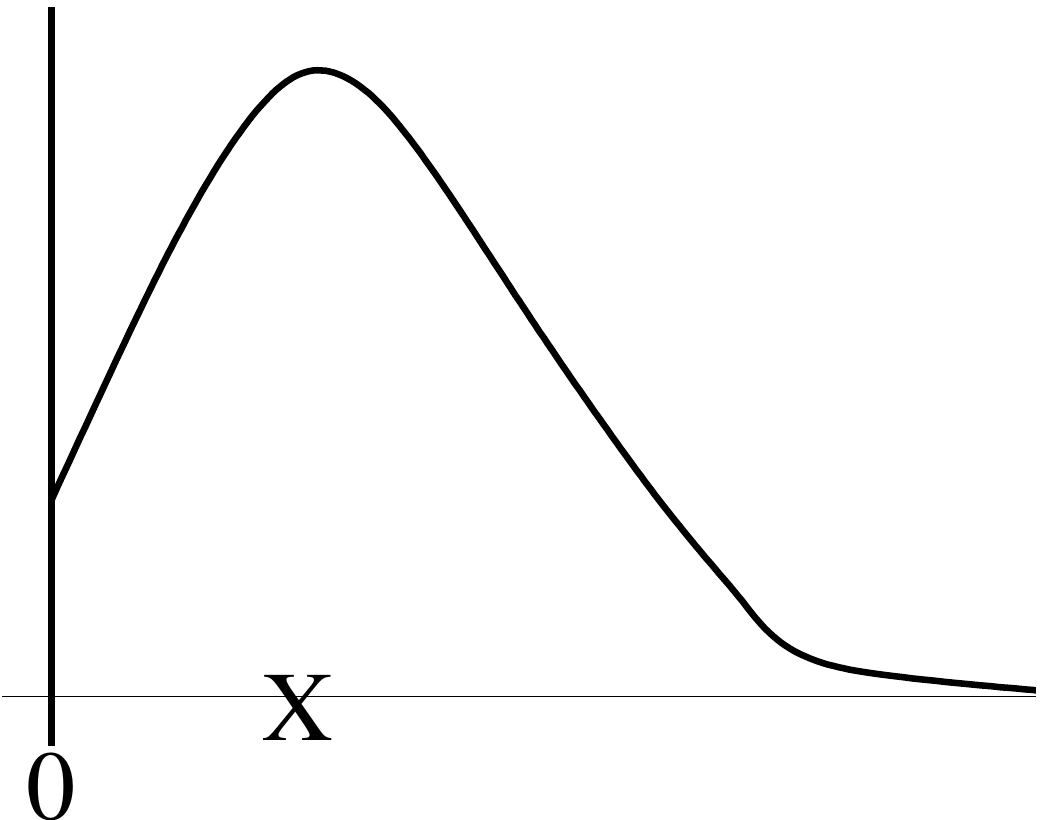}\\
\end{minipage}
\begin{minipage}{0.15\textwidth}
\scriptsize
$n=10^4$: 
\end{minipage}
\begin{minipage}{0.29\textwidth}
\includegraphics[width=0.7\textwidth,height=0.5\textwidth]{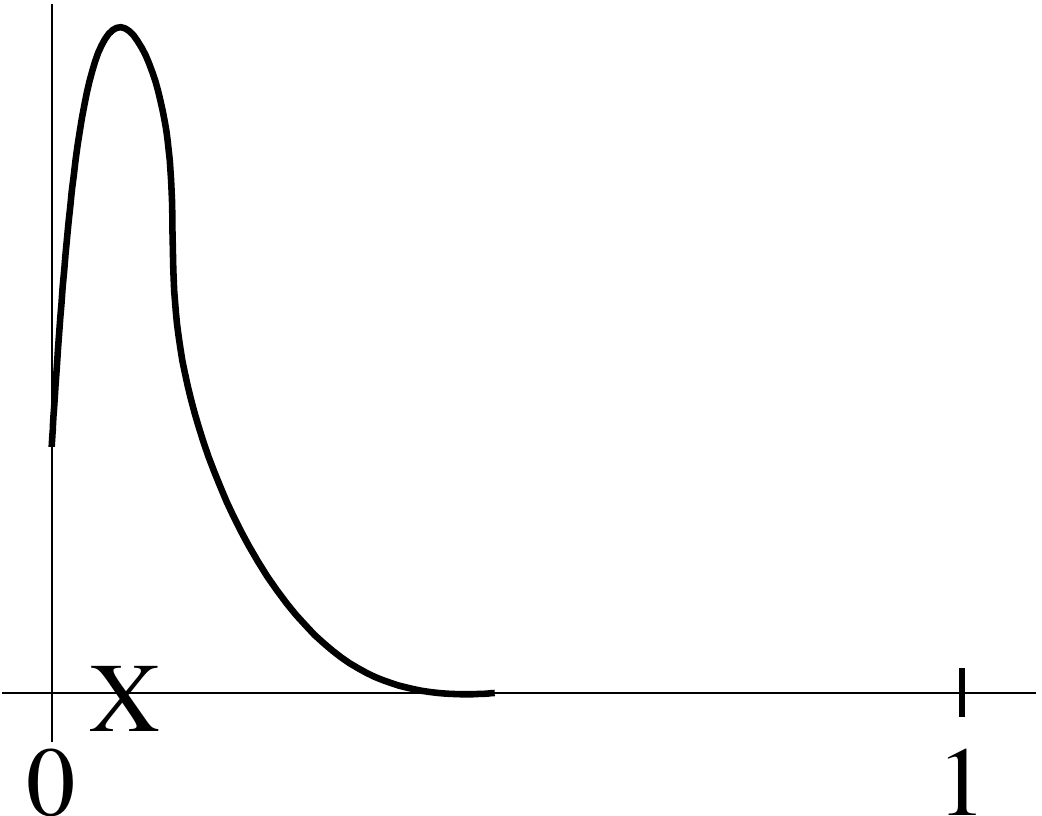}\\
\end{minipage}
\begin{minipage}{0.24\textwidth}
\small$\xrightarrow{\text{scale with } \sqrt{n}=100}$
\end{minipage}
\begin{minipage}{0.29\textwidth}
\includegraphics[width=0.7\textwidth,height=0.5\textwidth]{figures/scaled_distribution3.pdf}\\
\end{minipage}
\begin{minipage}{0.15\textwidth}
\centerline{$\downarrow$}
\end{minipage}
\begin{minipage}{0.29\textwidth}
\centerline{$\downarrow$}
\end{minipage}
\begin{minipage}{0.24\textwidth}
\mbox{}
\end{minipage}
\begin{minipage}{0.29\textwidth}
\centerline{$\downarrow$}
\end{minipage}
\begin{minipage}{0.15\textwidth}
\scriptsize 
$n=\infty$: 
\end{minipage}
\begin{minipage}{0.29\textwidth}
\includegraphics[width=0.7\textwidth,height=0.5\textwidth]{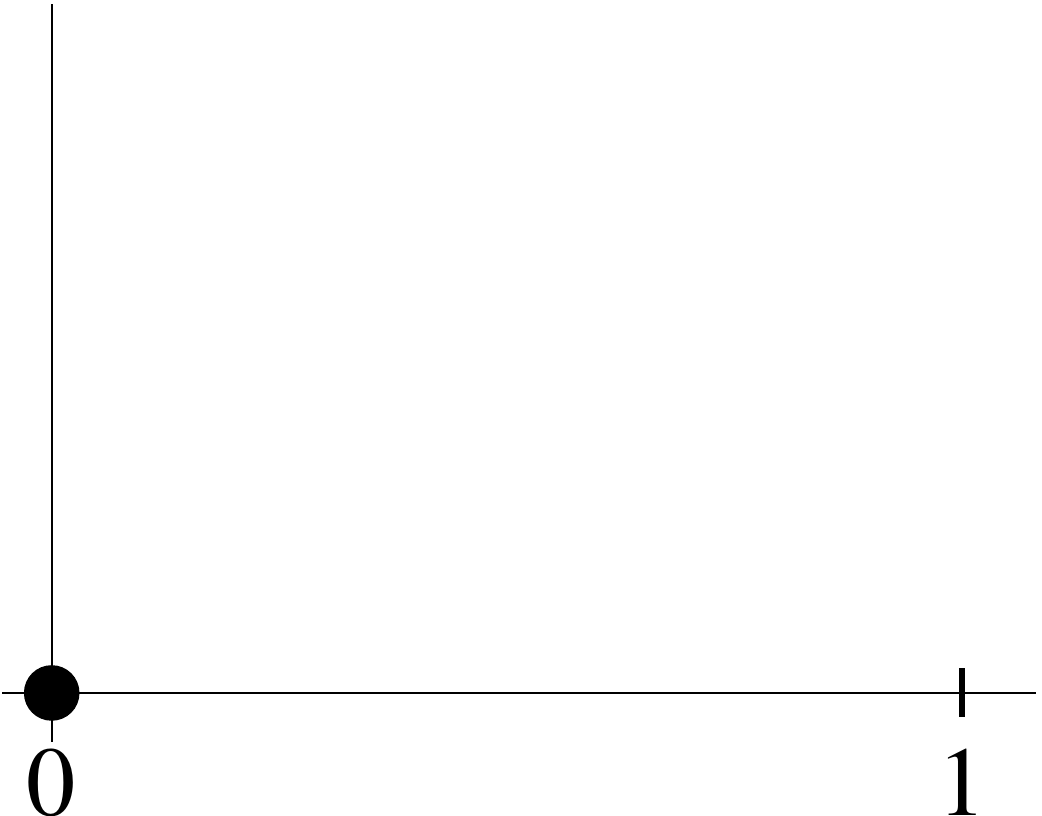}\\
\end{minipage}
\begin{minipage}{0.24\textwidth}
\mbox{}
\end{minipage}
\begin{minipage}{0.29\textwidth}
\includegraphics[width=0.7\textwidth,height=0.5\textwidth]{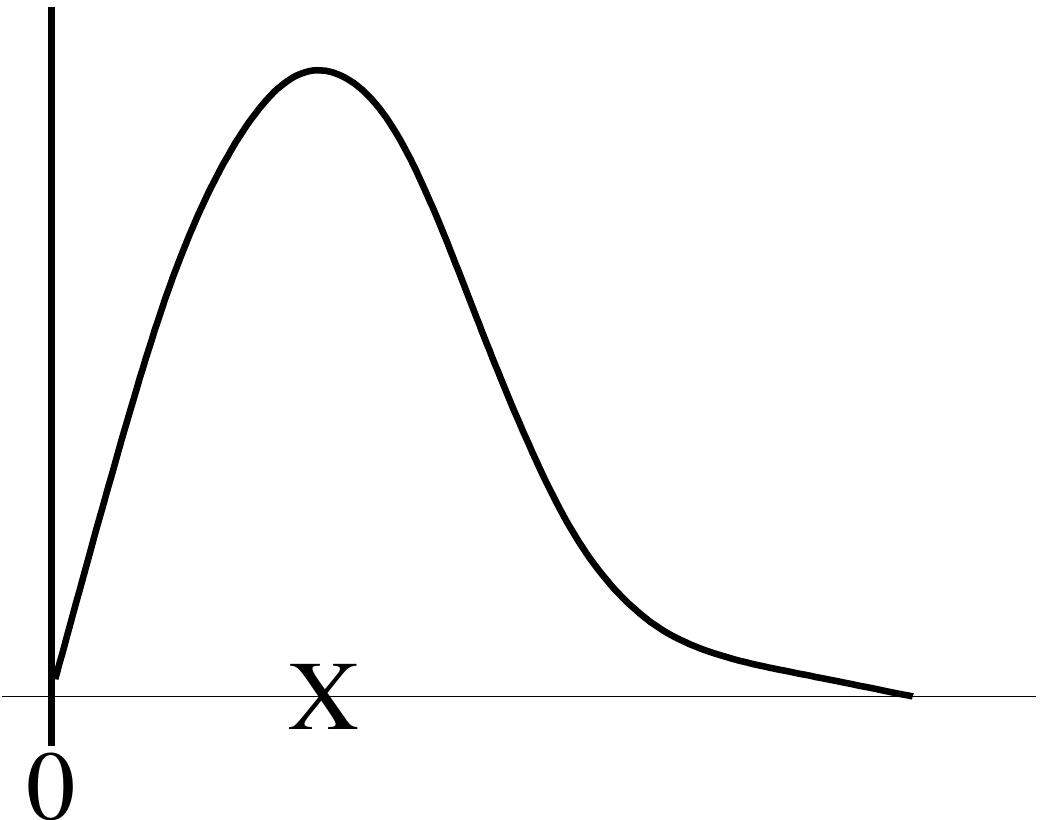}\\
\end{minipage}
}}
\end{center}
\caption{Different convergence processes. The left column shows the
  convergence studied in \th{th-convergence-simple}. 
As the sample size $n \to \infty$, the distribution of distances $\dmm(\Ccal, \Ccal')$
is degenerate, all mass is concentrated on 0. The right column shows
the convergence studied in \th{th-convergence-refined}. The
rescaled distances converge to a non-trivial distribution, and its
mean (depicted by the cross) is positive. To go from the left to the
right side one has to rescale by $\sqrt{n}$.
}
\label{fig-convergence}
\end{figure}

Note that even though the definition of instability in
\eq{def-instab-ohad} 
differs slightly from the definition in \eq{def-instab}, intuitively
it measures the same quantity. The definition in \eq{def-instab-ohad}
just has the technical advantage that all pairs of samples are
independent from one another. \\

{\em Proof sketch. } It is well known that if $Q^{(\infty)}_K$ has a unique global
minimum, then the centers constructed by the idealized $K$-means
algorithm on a finite sample satisfy a central limit theorem \citep{Pollard82}. That is,
if we rescale the distances between the sample-based centers and the
true centers with the factor $\sqrt{n}$, these rescaled distances
converges to a normal distribution as $n \to \infty$. When the cluster centers converge, the same can be
said about the cluster boundaries. In this case, instability
essentially counts how many points change side when the cluster
boundaries move by some small amount. The points that potentially
change side are the points close to the boundary of the true limit
clustering.  Counting these points is what the integrals $\int_{B_{ij}}
... p(x) dx$ in the definition of $\rinstab$ 
take care of.  The exact characterization of how the cluster
boundaries ``jitter'' can be derived from the central limit
theorem. This leads to the term $V_{ij} / \|c_i^{(*)} - c_j^{(*)}\|$ in the
integral.  $V_{ij}$ characterizes how the cluster centers themselves
``jitter''. The normalization $\|c_i^{(*)} - c_j^{(*)}\|$ is needed to transform
jittering of cluster centers to jittering of cluster boundaries: if
two cluster centers are very far apart from each other, the cluster
boundary only jitters by a small amount if the centers move by 
$\eps$, say. However, if the centers are very close to each other (say, they
have distance $3\eps$), then moving the centers by $\eps$ has a large
impact on the cluster boundary. The details of this proof are very
technical, we refer the interested reader to
\citealt{ShaTis08_colt,ShaTis09_nips}.  \ulesqed\\

Let us briefly
explain how the result in \th{th-convergence-refined} is compatible
with the result in \th{th-convergence-simple}.  On a high level, the
difference between both results resembles the difference between the
law of large numbers and the central limit theorem in probability
theory. The LLN studies the convergence of the mean of a sum of random
variables to its expectation (note that $\instab$ has the form of a
sum of random variables). The CLT is concerned with the same
expression, but rescaled with a factor $\sqrt{n}$. For the rescaled
sum, the CLT then gives results on the convergence in distribution.
Note that in the particular case of instability, the distribution of
distances lives on the non-negative numbers only. This is why the
rescaled instability in \th{th-convergence-refined} is positive and
not 0 as in the limit of $\instab$ in
Theorem~\ref{th-convergence-simple}. A toy figure explaining the
different convergence processes can be seen in \fig{fig-convergence}.\\

Theorem \ref{th-convergence-refined}  tells us that different parameters $k$
usually lead to different rescaled stabilities in the limit for $n \to
\infty$. Thus we can hope that if the sample size $n$ is large enough 
we can distinguish between different values of $k$ based on the
stability of the corresponding clusterings. An important question is
now which values of $k$ lead to stable and which ones lead to instable results,
for a given distribution $P$. \\

\subsection{Characterizing stable clusterings} \label{sec-characterization}

It is a straightforward consequence of \th{th-convergence-refined}
that if we consider different values $k_1$ and $k_2$ and the clustering objective
functions $Q^{(\infty)}_{k_1}$ and $Q^{(\infty)}_{k_2}$ have unique global minima, then the
rescaled stability values $\rinstab(k_1)$ and $\rinstab(k_2)$ can
differ from each other. Now we want to investigate which values of $k$ lead to
high stability and which ones lead to low stability. \\

\begin{conclusion}[Instable clusterings] \label{conclusion-instable}
  Assume that $Q^{(\infty)}_K$ has a unique global optimum.  If $\instab(K,n)$
  is large, the idealized $K$-means clustering tends to have
  cluster boundaries in high density regions of the space.

\end{conclusion}

There exist two different derivations of this conclusion, which have been
obtained independently from each other by completely different
methods \citep{BenLux08,ShaTis08_nips}. 
On a high level, the reason why the conclusion tends to hold is that
if cluster boundaries jitter in a region of high density, then more
points ``change side'' than if the boundaries jitter in a region of low
density.  \\

{\em First derivation, informal, based on
  \citet{ShaTis08_nips,ShaTis09_nips}. } Assume that $n$ is large
enough such that we are already in the asymptotic regime (that is,
the solution $\cn$ constructed on the finite sample is close to the
true population solution $\cstar$). Then the rescaled instability
computed on the sample is close to the expression given in
\eq{eq-rinstab}. If the cluster boundaries $B_{ij}$ lie in a high
density region of the space, then the integral in \eq{eq-rinstab} is
large --- compared to a situation where the cluster boundaries lie in
low density regions of the space. From a high level point of view,
this justifies the conclusion above. However, note that it is
difficult to identify how exactly the quantities $p$, $B_{ij}$ and
$V_{ij}$ influence $\rinstab$, as they are not independent of each other.\\

{\em Second derivation, more formal, based on \citet{BenLux08}}.  A
formal way to prove the conclusion is as follows. We introduce a new
distance $\dboundary$ between two clusterings. This distance measures
how far the cluster boundaries of two clusterings are apart from each
other. One can prove that the $K$-means quality function $Q^{(\infty)}_K$ is
continuous with respect to this distance function. This means that if
two clusterings $\Ccal, \Ccal'$ are close with respect to
$\dboundary$, then they have similar quality values. Moreover, if
$Q^{(\infty)}_K$ has a unique global optimum, we can invert this argument and
show that if a clustering $\Ccal$ is close to the optimal limit
clustering $\Ccal^*$, then the distance $\dboundary(\Ccal, \Ccal^*)$
is small.  Now consider the clustering $\Ccal^{(n)}$ based on a sample
of size $n$. One can prove the following key statement. If
$\Ccal^{(n)}$ converges uniformly (over the space of all probability
distributions) in the sense that with probability at least $1 -
\delta$ we have $\dboundary(\Ccal_n, \Ccal) \leq \gamma$, then 
\banum \label{eq-tube}
\instab(K,n) \leq 2 \delta + P(T_\gamma(B)). 
\eanum
Here $P(T_\gamma(B))$ denotes the probability mass of a tube of width
$\gamma$ around the cluster boundaries $B$ of $\Ccal$.  Results in
\citet{Bendavid07} establish the uniform convergence of the idealized
$K$-means algorithm.  This proves the conjecture: \eq{eq-tube}
shows that if $\instab$ is high, then there is a lot of mass
around the cluster boundaries, namely  the cluster boundaries are in
a region of high density.\\

For stable clusterings, the situation is not as simple. It is tempting
to make the following conjecture.

\begin{conjecture}[Stable clusterings] \label{conjecture-stable}
  Assume that $Q^{(\infty)}_K$ has a unique global optimum. If $\instab(K,n)$
  is ``small'', the idealized $K$-means clustering tends to have
  cluster boundaries in low density regions of the space.\\
\end{conjecture}

{\em Argument in favor of the conjecture: } As in the first approach above, considering the limit
expression of $\rinstab$ reveals that if the cluster boundary lies in
a low density area of the space, then the integral in $\rinstab$ tends to
have a low value.  In the extreme case where the cluster boundaries go
through a region of zero density, the rescaled instability is even 0.\\

{\em Argument against the conjecture: counter-examples! }  
One can construct artificial examples 
 where clusterings are stable although their
decision boundary lies in a high density region of the space (\citealp{BenLux08}). The way
to construct such examples is to ensure that the variations of the
cluster centers happen in parallel to cluster boundaries and not
orthogonal to cluster boundaries. In this case, the sampling variation
does not lead to jittering of the cluster boundary, hence the result
is rather stable. \\

These counter-examples show that Conjecture~\ref{conjecture-stable} cannot be true in
general. However, my personal opinion is that the counter-examples are
rather artificial, and that similar situations will rarely  be
encountered in practice. I believe that the conjecture ``tends to
hold'' in practice. It might be possible to formalize this intuition
by proving that the statement of the conjecture holds on a subset of ``nice'' and ``natural''
probability distributions. \\

The important consequence of Conclusion \ref{conclusion-instable} and
Conjecture~\ref{conjecture-stable} (if true) is the following.

\begin{conclusion}  \label{conclusion-idealized} 
{\bf (Stability of idealized $K$-means detects whether
  $K$ is too large)}
Assume that the underlying distribution $P$ has $K$ well-separated clusters,
and assume that these clusters can be represented by a center-based
clustering model. Then the following statements tend to hold for the idealized
$K$-means algorithm.  
\begin{enumerate} 
\item If $K$ is too large, then the clusterings obtained by the
  idealized $K$-means algorithm tend to be instable. 
\item If $K$ is correct or too small, then the clusterings obtained by the
  idealized $K$-means algorithm tend to be stable (unless the
  objective function has several global minima, for example due to
  symmetries). 
\end{enumerate}
\end{conclusion}

Given Conclusion \ref{conclusion-instable} and
Conjecture~\ref{conjecture-stable} it is easy to see why
Conclusion~\ref{conclusion-idealized} is true. If $K$ is larger than
the correct number of clusters, one necessarily has to split a true
cluster into several smaller clusters. The corresponding boundary goes
through a region of high density (the cluster which is being split).
According to Conclusion~\ref{conclusion-instable} this leads to
instability. If $K$ is correct, then the idealized (!) $K$-means
algorithm discovers the correct clustering and thus has decision
boundaries between the true clusters, that is in low density regions
of the space. If $K$ is too small, then the $K$-means algorithm has to
group clusters together. In this situation, the cluster boundaries are
still between true clusters, hence in a low density
region of the space.\\

\section{The actual $K$-means algorithm}  \label{sec-actual}

\begin{figure}[t]
\begin{center}
\includegraphics[width=0.8\textwidth]{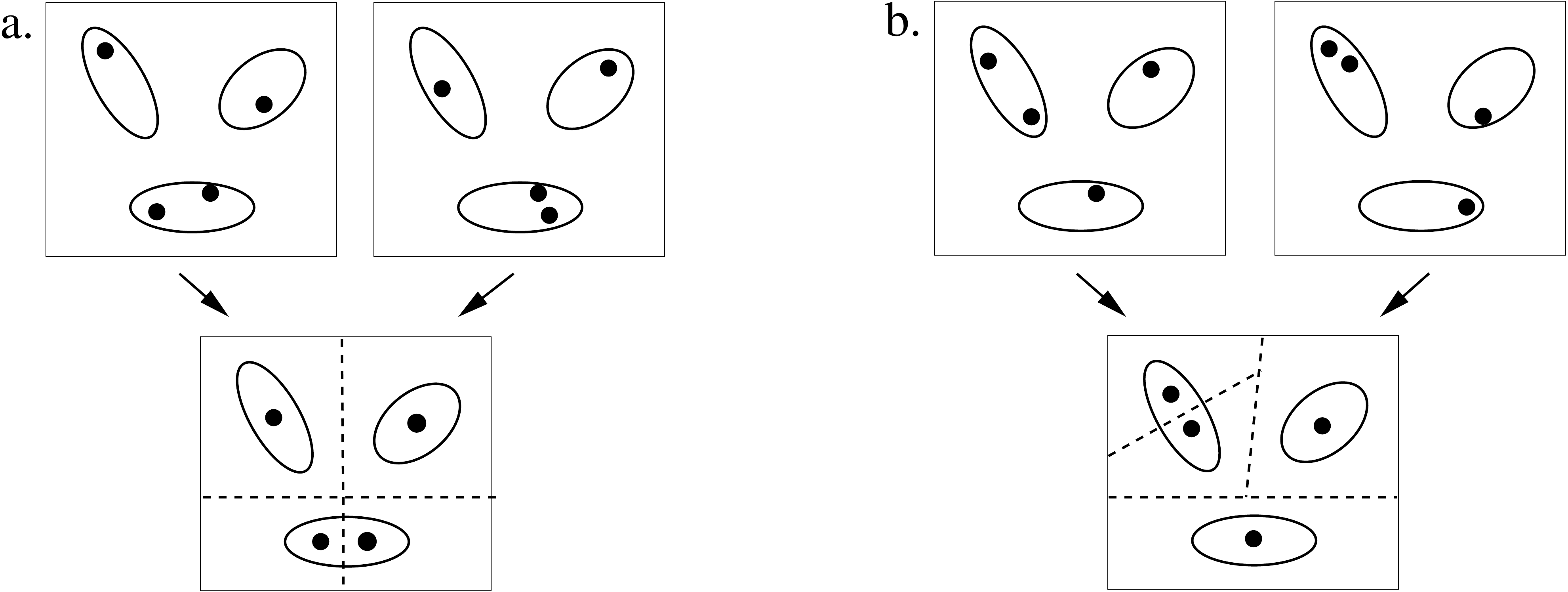}
\end{center}
\caption{Different initial configurations and the corresponding outcomes of the
  $K$-means algorithm. Figure a: the two boxes in the top row depict a data
  set with three clusters and four initial centers. Both boxes show
  different realizations of the same initial configuration. As can be
  seen in the bottom, both initializations lead to the same $K$-means
  clustering. Figure b: here the initial configuration is different
  from the one in Figure a, which leads to a different $K$-means
  clustering. }
\label{fig-init-configuration}
\end{figure}

In this section we want to study the actual $K$-means algorithm. In
particular, we want to investigate when and how it gets stuck in
different local optima. The general insight is that even though, from
an algorithmic point of view, it is an annoying property of the
$K$-means algorithm that it can get stuck in different local optima,
this property might
actually help us for the purpose of model selection. 
We now want to focus on the effect of the random
initialization of the $K$-means algorithm.  For simplicity, we ignore
sampling artifacts and assume that we always work with
``infinitely many'' data points;  that is, we work on the underlying
distribution directly. \\

The following observation is the key to our analysis.  Assume we are
given a data set with $\ktrue$ well-separated clusters, and assume
that we initialize the $K$-means algorithm with $\kinit \geq \ktrue$
initial centers.  The key observation is that if there is at least
one initial center in each of the underlying clusters, then {\em the
  initial centers tend to stay in the clusters they had been placed
  in. } This means that during the course of the $K$-means algorithm,
cluster centers are only  re-adjusted within the underlying
clusters and do not move between them. If this property is true,
then {the final clustering result is essentially determined by the
  {\em number} of initial centers in each of the true clusters. In
  particular, if we call the number of initial centers per cluster
  the {\em initial configuration}, one can say that each initial
  configuration leads to a unique clustering, and different
  configurations lead to different clusterings; see
  Figure~\ref{fig-init-configuration} for an illustration.  Thus, if
  the initialization method used in
  $K$-means regularly leads to different initial configurations, then we observe instability. \\

  In \citet{BubMeiLux09}, the first results in this direction were 
  proved. They are still preliminary in the sense that so far, proofs
  only exist for a simple setting.  However, we believe that the
  results also hold in a more general context.

\begin{theorem}[Stability of the actual $K$-means  algorithm]\label{th-actual-stability}
Assume that the underlying distribution $P$ is a mixture of two
well-separated Gaussians on $\R$. Denote the means of the Gaussians by
$\mu_1$ and $\mu_2$. 

\begin{enumerate}
\item 
Assume that we run the $K$-means algorithm with
$K=2$ and that we use an initialization scheme that places one initial center
in each of the true clusters (with high probability). Then the
$K$-means algorithm is stable in the sense that with high probability,
it  terminates in a solution with one center close to $\mu_1$ and one
center close to $\mu_2$.

\item 
Assume that we run the $K$-means algorithm with
$K=3$ and that we use an initialization scheme that places at least
one of the initial centers in each of the true clusters (with high probability). Then the
$K$-means algorithm is instable in the sense that with probability
close to 0.5 it terminates in a solution that considers the first Gaussian
as cluster, but splits the second Gaussian into two clusters; 
and with probability close to 0.5 it does it the other way round. 
\end{enumerate}

\end{theorem}

{\em Proof idea. }
The idea of this proof is best described with \fig{fig-proof-actual}. 
In the case of $\kinit=2$ one has to prove that if the one center lies in a
large region around $\mu_1$ and the second center in a similar region
around $\mu_2$, then the next step of $K$-means does not move the
centers out of their regions (in  \fig{fig-proof-actual}, these
regions are indicated by the black bars). If this is true, and if we know that there
is one initial center in each of the regions, the same is true when
the algorithm stops. Similarly, in the case of $\kinit=3$, one proves
that if there are two initial centers in the first region and one
initial center in the second region, then all centers stay in their
regions in one step of $K$-means. 
\ulesqed \\

\begin{figure}[t]
\begin{center}
\includegraphics[width=0.5\textwidth]{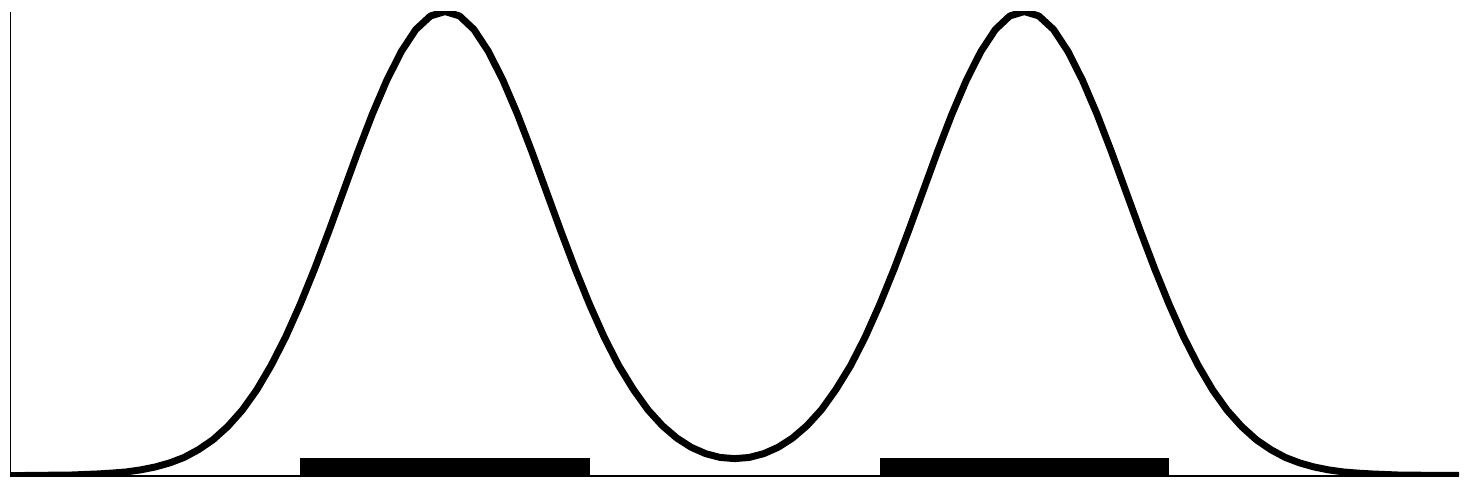}
\end{center}
\caption{Stable regions used in the proof of
  Theorem~\ref{th-actual-stability}. See text for details. }
\label{fig-proof-actual}
\end{figure}

All that is left to do now is to find an initialization scheme that
satisfies the conditions in
Theorem~\ref{th-actual-stability}. Luckily, we can adapt a scheme
that has already been used in \citet{DasSch07}. 
For simplicity, assume that all clusters have similar weights (for the
general case see \citealp{BubMeiLux09}), and that we want to
select $K$ initial centers for the
$K$-means algorithm. Then the following initialization should be used: \\

{\small\tt 
Initialization (I): 
\ulesquote{
\begin{enumerate}
\item Select $L$ preliminary centers uniformly at random
  from the given data set, where $L \approx K \log (K )$. 
\item Run one step of $K$-means, that is assign the data points to the
  preliminary  centers and re-adjust the centers once. 
\item Remove all centers for which the mass of the assigned data
  points is smaller than $p_0 \approx 1 / L$. 
\item Among the remaining centers, select $K$ centers by the
  following procedure: 
\begin{enumerate}
\item Choose the first center uniformly at random. 
\item Repeat until $K$ centers are selected: Select the next
  center as the one that maximizes the minimum distance to the
  centers  already selected. 
\end{enumerate}
\end{enumerate}
}
}

One can prove that this initialization scheme satisfies the conditions
needed in Theorem~\ref{th-actual-stability} (for exact details see
\citealp{BubMeiLux09}).

\begin{theorem}[Initialization]\label{th-init} 
  Assume we are given a mixture of $\ktrue$ well-separated Gaussians
  in $\R$, and denote the centers of the Gaussians by $\mu_i$. If we
  use the Initialization (I) to select $\kinit$
  centers, then there
exist $\ktrue$ disjoint regions ${A}_k$ with
$\mu_k\in {A}_k$, so that
 all $\kinit$ centers are contained in one of the $A_k$ and 
\begin{itemize}
\item if $\kinit=\ktrue$, each $A_k$ contains exactly one center, 
\item if $\kinit<\ktrue$, each $A_k$  contains at most one center, 
\item if $\kinit>\ktrue$, each $A_k$ contains at least one center. 
\end{itemize}
\end{theorem}

{\em Proof sketch. } The following statements can be proved to hold
with high probability. By selecting $\ktrue \log(\ktrue)$ preliminary
centers, each of the Gaussians receives at least one of these
centers. By running one step
of $K$-means and removing the
centers with too small mass, one removes all preliminary centers that
sit on outliers. Moreover, one can prove that ``ambiguous centers''
(that is, centers that sit between two clusters) attract only few
data points  and will be removed  as well. Next one
shows that centers that are ``unambiguous'' are reasonably close to a
true cluster center $\mu_k$. Consequently, the method for selecting
the final center from the remaining preliminary ones ``cycles though
different Gaussians'' before visiting a particular Gaussian for the second time. 
\ulesqed \\

\begin{figure}[t]
\begin{center}
\includegraphics[width=0.3\textwidth]{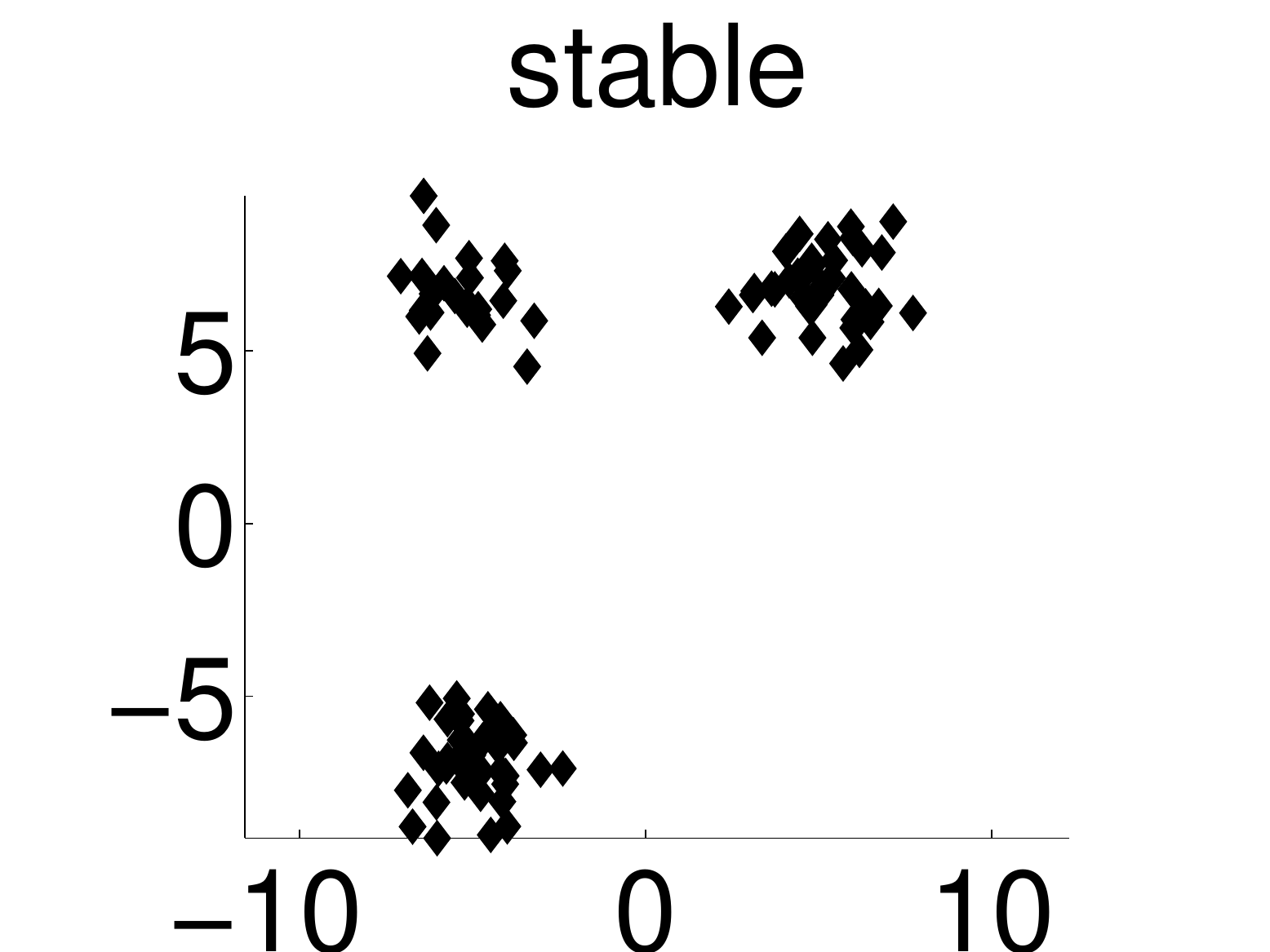}\hspace{1cm}
\includegraphics[width=0.3\textwidth]{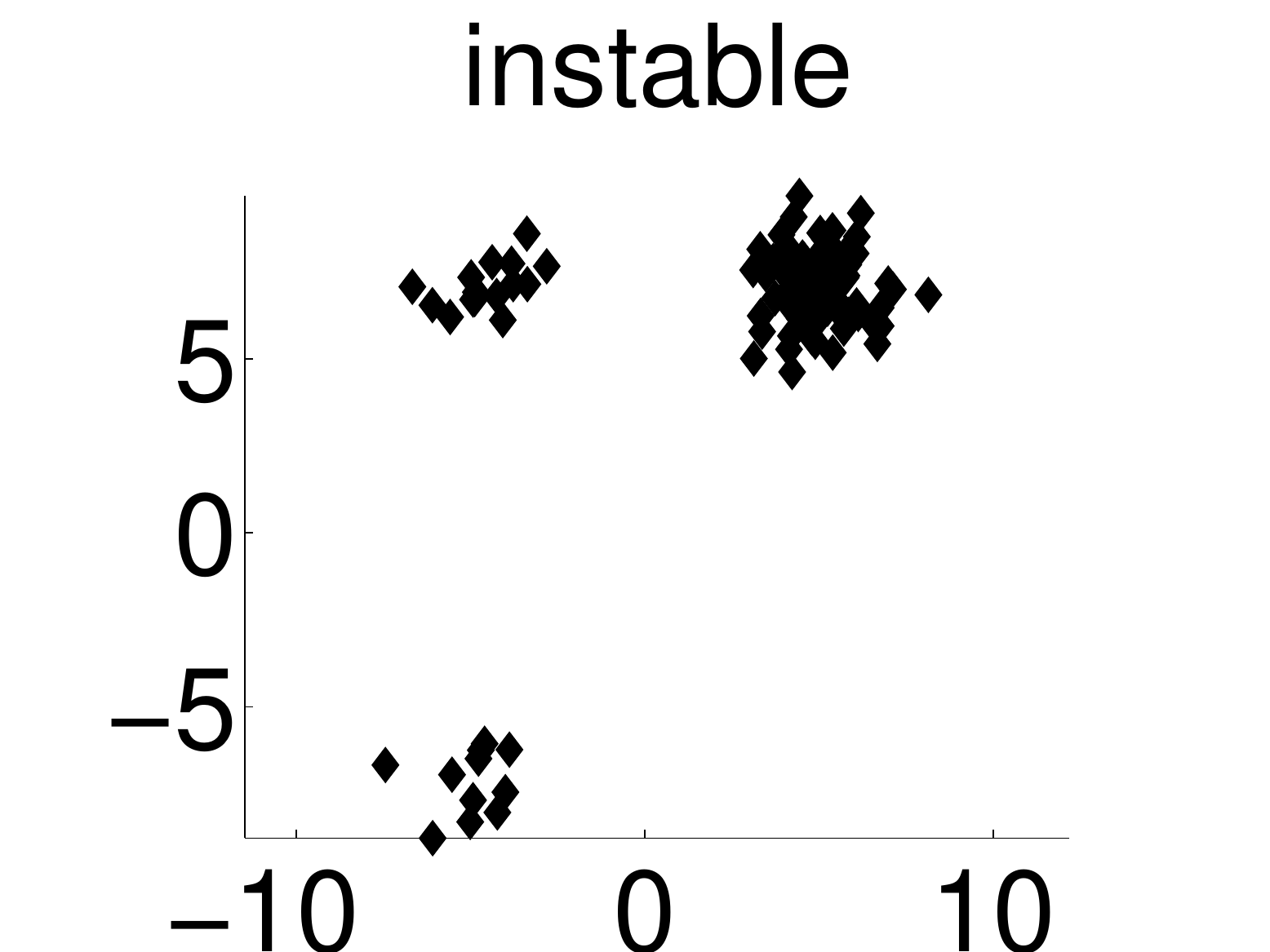}
\end{center}
\caption{Illustration for the case where $K$ is too small. 
We consider two data sets that have been drawn from a mixture of three
  Gaussians with means 
$\mu_1 = (-5,-7)$, 
$\mu_2 = (-5,7)$, 
$\mu_3 = (5,7)$ and unit variances. In the left figure, all clusters
have the same weight $1/3$, whereas in the right figure the top right
cluster has larger weight 0.6 than the other two clusters with weights
0.2 each. If we run $K$-means with $K=2$, we can verify experimentally
that the algorithm is pretty stable if applied to points from the distribution in the
left figure. It nearly always merges the top two
clusters. On the distribution shown in the right figure, however, the algorithm is
instable. Sometimes the top two clusters are merged, and sometimes the
left two clusters. }
\label{fig-too-small}
\end{figure}

When combined, the results of Theorems~\ref{th-actual-stability} and
\ref{th-init} show that if the data set contains $\ktrue$
well-separated clusters, then the $K$-means algorithm is stable if it
is started with the true number of clusters, and instable if the
number of clusters is too large. Unfortunately, in the case where $K$
is too small one cannot make any useful statement about stability
because the aforementioned  configuration argument  does not hold any
more. In particular, initial cluster centers do not
stay inside their initial clusters, but move out of
the clusters. Often, the final centers constructed by the $K$-means
algorithm lie in between several true clusters, and it is very hard to
predict the final positions of the centers from the initial ones.  This can be seen
with the  example shown in Figure \ref{fig-too-small}. We
consider two data sets from a mixture of three Gaussians. The only
difference between the two data sets  is that in the left plot all mixture
components have the same weight, while in the right plot the top right
component has a larger weight than the other two components. One can
verify experimentally that if initialized with $\kinit = 2$, the
$K$-means algorithm is rather stable in the left figure (it always
merges the top two clusters). But it is instable in the right figure
(sometimes it merges the top clusters, sometimes the left two
clusters). This example illustrates that if the number of clusters is
too small, subtle differences in the distribution can decide on
stability or instability of the actual $K$-means algorithm. \\

In general, we expect that the following statements hold
(but they have not yet been proved in a context more general than in 
Theorems~\ref{th-actual-stability} and \ref{th-init}). \\

\begin{conjecture}[Stability of the actual $K$-means algorithm] \label{conjecture-stable-actual}
Assume that the underlying distribution has $\ktrue$ well-separated
clusters, and that these clusters can be represented by a center-based
clustering model. Then, if one uses Initialization (I)  to
construct $\kinit$ initial centers, the
following statements hold: 
\blobb{If $\kinit = \ktrue$, we have one center per cluster, with
  high probability. The clustering results are stable.}
\blobb{If $\kinit > \ktrue$, different initial configurations
  occur. By the above argument, different configurations lead to different
  clusterings, so we observe instability. }
\blobb{If $\kinit < \ktrue$, then depending on subtle differences in
  the underlying distribution we can have 
 either stability or instability. \\}
\end{conjecture}

\section{Relationships between the results}  \label{sec-relationships}

In this section we discuss conceptual aspects of the results and relate them to each other.

\subsection{Jittering versus jumping}

\begin{figure}[t]
\begin{center}
\includegraphics[width=0.7\textwidth]{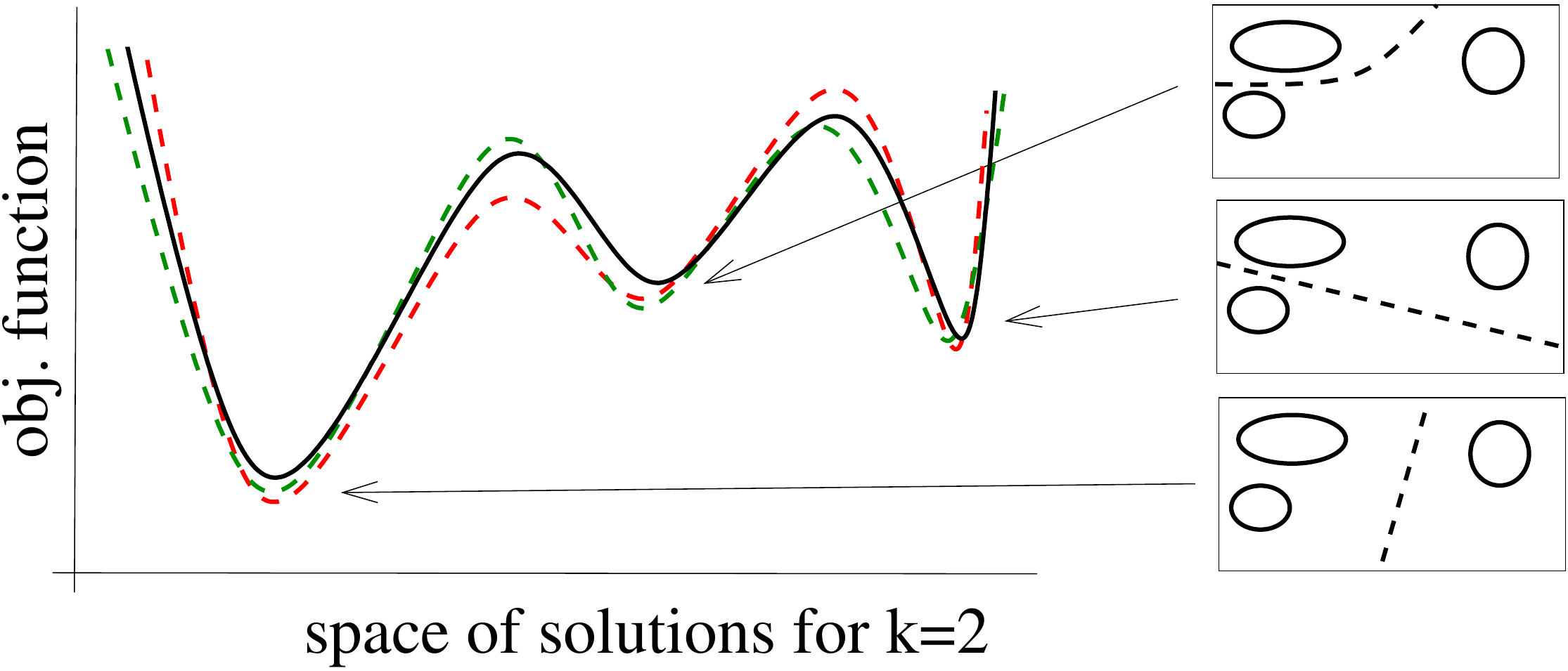}
\end{center}
\caption{The $x$-axis depicts the space of all clusterings for a fixed
  distribution $P$ and for a fixed parameter $K$ (this is an
  abstract sketch only). The $y$-axis shows the value
of the objective function of the different solutions. The solid line
corresponds to the true limit objective function $Q^{(\infty)}_K$, the dotted lines
show the sample-based function $Q^{(\infty)}_K$ on different samples. The
idealized $K$-means algorithm only studies the jittering of the global
optimum, that is how far the global optimum varies due to the sampling
process. The jumping between different local optima is induced by
different random initializations, as investigated for the actual
$K$-means algorithm.  }
\label{fig-jittering}
\end{figure}

There are two main effects that 
lead to instability of the $K$-means algorithm. Both effects
are visualized in \fig{fig-jittering}. \\

{\em Jittering of the cluster boundaries. } Consider a fixed local (or
global) optimum of $Q^{(\infty)}_K$ and the corresponding clustering on different random
samples.  Due to the fact that different samples lead to slightly
different positions of the cluster centers, the cluster boundaries
``jitter''. That is, the cluster boundaries corresponding to different
samples are slightly shifted with respect to one another.  We call this behavior the ``jittering'' of a
particular clustering solution. For the special case of the global
optimum, this jittering has been investigated in Sections
\ref{sec-convergence-refined} and \ref{sec-characterization}.  It has
been established that different parameters $K$ lead to different
amounts of jittering (measured in terms of rescaled instability). The
jittering is larger if the cluster boundaries are in a high density
region and smaller if the cluster boundaries are in low density
regions of the space. The main ``source'' of jittering is the
sampling variation. \\

{\em Jumping between different local optima. } By ``jumping'' we refer
to the fact that an algorithm terminates in different local
optima. Investigating jumping has been the major goal in Section
\ref{sec-actual}.  The main source of jumping is the
random initialization. If we initialize the $K$-means algorithm in
different configurations, we end in different local optima.  The key
point in favor of clustering stability is that one can relate the
number of local optima of $Q^{(\infty)}_K$ to whether the number $K$ of
clusters is correct or too large (this has happened implicitly in
Section~\ref{sec-actual}).

\subsection{Discussion of the main theorems} 

{\em Theorem~\ref{th-convergence-simple}} works in the idealized
setting. In Part 1 it shows that if the underlying distribution is not
symmetric, the idealized clustering results are stable in the sense
that %
different samples always lead
to the same clustering. That is, no jumping between different
solutions takes place. In hindsight, this result can be considered as an
artifact of the idealized clustering scenario. The idealized $K$-means
algorithm artificially excludes the possibility of ending in different
local optima. Unless there exist several global optima, jumping
between different solutions cannot happen. In particular, the
conclusion that clustering results are stable for all values of $K$ does not carry over
to the realistic $K$-means
algorithm (as can be seen from the results in \sec{sec-actual}). Put
plainly, even though the idealized $K$-means algorithm with $K=2$ is
stable in the example of \fig{fig-stability-wrong}a, the actual $K$-means algorithm
is instable. \\
Part 2 of Theorem~\ref{th-convergence-simple} states that if the
objective function has several global optima, for example due to
symmetry, then jumping takes place even for the idealized $K$-means
algorithm and results in instability. In the setting of the theorem,
the jumping is merely induced by having different random
samples. However, a similar result can be shown to hold for the actual
$K$-means algorithm, where it is induced due to random
initialization. Namely, if the underlying distribution is perfectly
symmetric, then ``symmetric initializations'' lead to the different
local optima corresponding to the different symmetric solutions.\\

To summarize, Theorem~\ref{th-convergence-simple} investigates whether
jumping between different solutions takes place due to the random
sampling process.
The
negative connotation of Part 1 is an artifact of the idealized setting
 that  does not carry over to the actual $K$-means algorithm, whereas the
positive connotation of Part 2 does carry
over. \\

{\em Theorem~\ref{th-convergence-refined}} studies how
different samples affect the jittering of a unique solution of the
idealized $K$-means algorithm. In general, one can expect that
similar jittering takes place for the actual $K$-means algorithm as
well. In this sense, we believe that the results of this theorem can
be carried over to the actual $K$-means algorithm. \\
However, if we reconsider the intuition stated in the introduction and
depicted in \fig{fig-stability-idea}, we realize that
jittering was not really what we had been looking for. The main intuition
in the beginning was that the algorithm might jump between different
solutions, and that such jumping shows that the underlying parameter
$K$ 
is wrong.  In practice, stability is usually computed for the actual
$K$-means algorithm with random initialization and on different
samples. Here both effects (jittering and jumping) and both random
processes (random samples and random initialization) play a role. We
suspect that the effect of jumping to different local optima due to
different initialization has higher impact on stability than the
jittering of a particular solution due to sampling variation. Our
reason to believe so is that the distance between two clusterings is
usually higher if the two clusterings correspond to different local
optima than if they correspond to the same solution with a slightly
shifted boundary. \\

To summarize, Theorem~\ref{th-convergence-refined} describes the jittering
behavior of an individual solution of the idealized $K$-means
algorithm. We believe that similar effects take place for the actual
$K$-means algorithm. However, we also believe that the influence of
jittering on stability plays a minor role compared to the one of jumping. \\

{\em Theorem~\ref{th-actual-stability}} investigates
the jumping behavior of the actual $K$-means algorithm. As the source of
jumping, it considers the random initialization only. It does not take
into account variations due to random samples (this is hidden in the
proof, which works on the underlying distribution rather than with
finitely many sample points). However, we believe that the
results of this theorem also hold for finite
samples. Theorem~\ref{th-actual-stability} is not yet as general as we
would like it to be. But we believe that studying the jumping
behavior of the actual $K$-means algorithm is the key to understanding 
the stability of the $K$-means algorithm used
in practice, and Theorem~\ref{th-actual-stability} points in the right
direction. \\

{\em Altogether, } the results obtained in the idealized and realistic setting
perfectly complement each other and describe two sides of the same coin. The
idealized setting mainly studies what influence the different samples
can have on the stability of one particular solution.  The realistic
setting focuses on how the random initialization makes the algorithm
jump between different local optima. In both settings, stability
``pushes'' in the same direction: If the number of clusters is too
large, results tend to be instable. If the number of clusters is
correct, results tend to be stable. If the number of clusters is too
small, both stability and instability can occur, depending on subtle
properties of the underlying distribution. \\

\chapter{Beyond $K$-means}  \label{sec-beyond}

Most of the theoretical results in the literature on clustering stability have been
proved with the $K$-means algorithm in mind. However, some of them hold for
more general clustering algorithms. This is mainly the case for
 the idealized clustering setting. \\

Assume a general clustering objective function $Q$ and an ideal
clustering algorithm that globally minimizes this objective
function. If this clustering algorithm is consistent in the sense that
the optimal clustering on the finite sample converges to the optimal
clustering of the underlying space, then the results of Theorem
\ref{th-convergence-simple} can be carried over to this general
objective function \citep{BenLuxPal06}.  Namely, if the objective
function has a unique global optimum, the clustering algorithm is
stable, and it is instable if the algorithm has several global minima
(for example due to symmetry).  It is not too surprising that one can
extend the stability results of the $K$-means algorithm to more
general vector-quantization-type algorithms. However, the setup of
this theorem is so general that it also holds for completely different
algorithms such as spectral clustering.  The consistency requirement
sounds like a rather strong assumption. But note that clustering algorithms that are
not consistent are completely unreliable and should not be used
anyway. \\

Similarly as above, one can also generalize the characterization of
instable clusterings stated in Conclusion \ref{conclusion-instable}, 
cf. \citet{BenLux08}. Again we are dealing with algorithms that minimize an
objective function. The  consistency requirements are slightly
stronger in that we need uniform consistency over the space (or a
subspace) of probability distributions. Once such uniform consistency
holds, the characterization that instable clusterings tend to
have their boundary in high density regions of the space can be
established. \\

While the two results mentioned above can be carried over to a huge
bulk of clustering algorithms, it is not as simple for the refined
convergence analysis of Theorem \ref{th-convergence-refined}. Here we
need to make one crucial additional assumption, namely the existence
of a central limit type result. This is a rather strong assumption
which is not satisfied for many clustering objective
functions. However, a few results can be established
\citep{ShaTis09_nips}: in addition to the traditional $K$-means
objective function, a central limit theorem can be proved for other
variants of $K$-means such as kernel $K$-means (a kernelized version
of the traditional $K$-means algorithm) or Bregman divergence
clustering (where one selects a set of centroids such that the average
divergence between points and centroids is minimized). Moreover,
central limit theorems are known for maximum likelihood
estimators, which leads to stability results for certain types of
model-based clusterings using maximum likelihood
estimators. Still  the results of Theorem
\ref{th-convergence-refined} are limited to a small number of clustering
objective functions, and one cannot expect to be able to
extend them to a wide range of clustering algorithms. \\

Even stronger limitations hold for the results about the actual
$K$-means algorithm. The methods used in Section \ref{sec-actual} were
particularly designed for the $K$-means algorithm. It might be
possible to extend them to more general centroid-based algorithms, but
it is not obvious how to advance further.  In spite of this shortcoming, we
believe that these results hold in a much more general context of
randomized clustering algorithms. 
From a high level point of view, the actual $K$-means algorithm is a
randomized algorithm due to its random initialization. The
randomization is used to explore different local optima of the
objective function. There were two key insights in our stability
analysis of the actual $K$-means algorithm: First, we could describe
the ``regions of attraction'' of different local minima, that is we
could prove which initial centers lead to which solution in the end
(this was the configurations idea). 
Second, we could relate the ``size'' of the
regions of attraction to the number of clusters. %
Namely, if the number of
clusters is correct, the global minimum will have a huge region of
attraction in the sense that it is very likely that we will end in the
global minimum. If the number of clusters is too large, we could show
that there exist several local optima with large regions of
attraction. This leads to a significant likelihood of
ending in different local optima and observing instability.\\

We believe that similar arguments can be used to investigate stability
of other kinds of randomized clustering algorithms. However, such an
analysis always has to be adapted to the particular algorithm under
consideration. In particular, it is not obvious whether the number of
clusters can always be related to the number of large regions of
attraction. Hence it is  an open question whether 
results similar to the ones for the actual $K$-means algorithm also hold for completely
different randomized clustering algorithms. \\

\chapter{Outlook}  \label{sec-outlook}

Based on the results presented above one can draw a cautiously 
optimistic picture about model selection based on clustering
stability for the $K$-means algorithm. Stability can
discriminate between different values of $K$, and the values of $K$
that lead to stable results have desirable properties. If the
data set contains a few well-separated clusters that can be
represented by a center-based clustering, then stability has the
potential to discover the correct number of clusters. \\

An important point to stress is that stability-based model selection
for the $K$-means algorithm can only lead to convincing  results if the
underlying distribution can be represented by center-based
clusters. If the clusters are very elongated or have complicated
shapes, the $K$-means algorithm cannot find a good representation of
this data set, regardless what number $K$ one uses. In this case,
stability-based model selection breaks down, too. It is a
legitimate question what implications this has in practice. We
usually do not know whether a given data set can be represented
by center-based clusterings, and often the $K$-means algorithm is
used anyway. In my opinion, however, the question of selecting the
``correct'' number of clusters is not so important in this case. The
only way in which complicated structure can be represented using
$K$-means is to break each true cluster in several small, spherical
clusters and either live with the fact that the true clusters are split in
pieces, or use some mechanism to join these pieces afterwards to form
a bigger cluster of general shape. In
such a scenario it is not so important what number of
clusters we use in the $K$-means step: it does not really matter whether we split an
underlying cluster into, say,  5 or 7 pieces.\\

There are a few technical questions that deserve further
consideration. Obviously, the results in \sec{sec-actual} are still
somewhat preliminary and should be worked out in more generality.  The
results in Section~\ref{sec-idealized} are large sample results. It is 
not clear what ``large sample size'' means in practice, and one can
construct examples where the sample size has to be arbitrarily large
to make valid statements \citep{BenLux08}. However, such examples can
either be countered by introducing assumptions on the underlying
probability distribution, or one can state that the sample size has to
be large enough to ensure that the cluster structure is
well-represented in the data and that we don't miss any clusters. \\

There is yet another limitation that is more severe, namely
the number of clusters to which the results apply.  The conclusions in
\sec{sec-idealized} as well as the results in \sec{sec-actual} only
hold if the true number of clusters is relatively small (say, on the
order of 10 rather than on the order of 100), and if the parameter $K$
used by $K$-means is in the same order of magnitude. Let us briefly
explain why this is the case. 
In the idealized setting, the limit results in Theorems \ref{th-convergence-simple} and
\ref{th-convergence-refined} of course hold regardless of what the true
number of clusters is. But the subsequent interpretation regarding
cluster boundaries in high and low density areas breaks down if the
number of clusters is too large.  The reason is that the influence of
one tiny bit of cluster boundary between two clusters is negligible
compared to the rest of the cluster boundary if there are many
clusters, such that other factors might dominate the behavior of
clustering stability. 
In the realistic setting of
Section~\ref{sec-actual}, we use an initialization scheme
which, with high probability, places centers in different clusters
before placing them into the same cluster. The procedure works well if
the number of clusters is small. However, the larger the number of
clusters, the higher the likelihood to fail with this
scheme. Similarly problematic is the situation where the true number of
clusters is small, but the $K$-means algorithm is run with a very
large $K$. 
Finally, note that similar limitations hold for all model selection
criteria. It is simply a very difficult (and pretty useless) question 
whether
a data set contains 100 or 105 clusters, say. \\

While stability is relatively well-studied for the $K$-means
algorithm, there does not exist much work on the stability of
completely different clustering mechanisms. We have seen in Section
\ref{sec-beyond} that some of the results for the idealized $K$-means
algorithm also hold in a more general context. However, this is not
the case for the results about the actual $K$-means algorithm. We
consider the results about the actual $K$-means algorithm as the
strongest evidence in favor of stability-based model selection for
$K$-means. Whether this principle can be proved to work well for
algorithms very different from $K$-means is an open question.\\

An important point we have not discussed in depth is how clustering
stability should be implemented in practice. As we have outlined in
Section \ref{sec-implementation} there exist many different protocols
for computing stability scores. It would be very important to compare
and evaluate all these approaches in practice, in particular as there
are several unresolved issues (such as the
normalization). Unfortunately, a thorough study that
compares all different protocols in practice does not
exist. \\

\bibliography{general_bib,ules_publications,ules_publications_submitted}

\end{document}